\documentclass[cameraready]{Interspeech}

\usepackage{makecell}
\usepackage{xcolor}
\usepackage{soul}
\usepackage{hyperref}
\usepackage{subfig}
\usepackage{balance}
\usepackage{graphicx}
\usepackage{url}
\newcommand{\HL}[1]{\hl{#1}}
\renewcommand{\HL}[1]{#1}
\newcommand{\highlight}[1]{\hl{#1}}
\renewcommand{\highlight}[1]{#1}

\title{Dialogue to Detection: A Multimodal Hybrid NLP Pipeline for Insurance Fraud Detection}

\author[affiliation={1}, orcid=0000-0001-6813-5372, correspondingauthor]{Muhammad Shakeel}{Akram}
\author[affiliation={1}, orcid=0000-0003-4647-9996]{Amal}{Htait}
\author[affiliation={1}, orcid=0000-0002-9825-59111]{Abdul Hamid}{Sadka}
\author[affiliation={2}]{Emma}{Meisingseth}
\author[affiliation={2}, orcid=0009-0007-4747-8764]{Karishma}{Jaitly}

\address{
    $^1$ Aston University, Birmingham,UK. 
    $^2$ Domestic \& General, Wimbledon, UK.
}
\email{$^1$\{m.akram5, a.htait, a.sadka\}@aston.ac.uk, \\ $^2$\{Emma.Meisingseth, Karishma.Jaitly\}@domesticandgeneral.com}


\usepackage{comment}

\begin{document}

\maketitle

\abstract{
Insurance fraud imposes substantial financial losses and operational inefficiencies, raising premiums and impacting trust among legitimate policyholders. Early detection at FNOL remains a persistent challenge. Existing approaches rely largely on private, text-only datasets, limiting progress on multimodal methods that integrate linguistic, behavioural, and speaker-based indicators. We introduce a synthetic multimodal framework that replicates FNOL conditions. It generates agent–customer dialogue transcripts and two-speaker audios, performs ASR and diarisation. Downstream modules combine NER, regex-based feature extraction, LLM–RAG retrieval, and speaker embeddings in a rule-based risk score to flag narrative reuse, structural inconsistencies, and cross-case voice repetition while balancing sensitivity and false positives. Dataset validation and component-level evaluations show stability and transfer potential, offering a reproducible baseline beyond text-only fraud detection.

}

\section{Introduction}
Insurance fraud is a persistent and high-cost problem worldwide, with estimates putting annual losses over \$300 billion in the US and more than £1.1 billion in the UK \cite{govuk2024fraud,ifb2024report,naic2024fraud,aslam2022insurance,ali2022financial}. Such losses affect not only insurers, but also legitimate policyholders through increased premiums and reduced benefits. The claim process, especially at the FNOL stage, is dominated by narrative data: spoken reports through call centres and written submissions through web portals. Detecting fraud in this early phase is both strategically advantageous and operationally challenging.

Traditional fraud detection relies on manual review/static rule-based systems, which are prone to human error and inefficient for large-scale \cite{backlund2023detection}. With the rise of AI and NLP, automated fraud detection from customer call transcripts has become a viable alternative. Transformer-based models such as BERT, GPT, and XLNet \cite{devlin2019bert,liu2024gpt,yang2019xlnet} have revolutionised NLP, and when fine-tuned on insurance-specific data, can classify and interpret claim narratives with high~accuracy.
However, progress remains constrained by the lack of publicly available \emph{multimodal} datasets containing paired speech and text data from insurance claims. Proprietary datasets exist within industry but are inaccessible due to privacy regulations (GDPR)
and contractual constraints \cite{piehl2021classification,backlund2023detection}. Unlike public speech corpora such as LibriSpeech or Switchboard, no openly licensed dataset provides insurance claim agent–customer dialogues and recordings with fraud labels. Existing call-centre datasets are often tabular or text-only and largely drawn from non-insurance domains 
\cite{mendeley2020insurance,aixblock2024calls}. Yet speech carries paralinguistic deception cues, such as tone, hesitation, and speaker traits \cite{leal2018telephone,bajaj2020fraud}, while text enables scalable entity extraction and semantic retrieval. Integrating both modalities allows cross-verification of voice identity and narrative similarity to uncover organised fraud clusters. Consequently, research consistently identifies data scarcity as the central bottleneck \cite{ali2022financial,aslam2022insurance,banulescu2024practical}.

Given the absence of shareable real-world datasets and the need for multimodal evidence, we \HL{introduce an end-to-end \emph{synthetic data generation and analysis pipeline} to simulate FNOL conditions while preserving realistic entity distributions, conversational patterns, and audio characteristics.} The workflow uses GPT-2 to generate legitimate and fraudulent dialogues with embedded entities; xTTS to synthesise two-speaker audio; WhisperX for STT (Speech-to-Text using ASR) and diarisation; NER+Regex for entity extraction; Resemblyzer embeddings and cosine similarity for cross-case voice reuse detection; and BERT with RAG for historical narrative matching. The output of each stage is then fed into a rule-based scoring system that estimates the risk of fraud. This design aims to unlock research without access to sensitive calls while staying faithful to operational constraints (two-channel audio, FNOL text summaries, diarisation errors, incomplete histories) \HL{and establishes a system architecture that enables reproducible evaluation and provides a foundation for future~extensions.}

The key contributions of this work are:

\begin{itemize}
  \item We design a reproducible \emph{synthetic multimodal FNOL dataset}, 
  balancing fraudulent and legitimate claims, while preserving realistic distributions of entities, dialogue structures, and audio properties.
  \item \HL{We introduce the first end-to-end \emph{multimodal pipeline for insurance fraud analysis}, spanning synthetic dialogue generation, multi-speaker audio synthesis, ASR+diarisation, and} combining semantic retrieval and entity extraction from text with speaker-embedding similarity to surface repeated narratives and voice reuse across claims.
  \item We propose a \emph{transparent risk-scoring framework} that fuses structured, vocal, and textual signals into an interpretable low to high fraud scale, enabling the pipeline to be a reproducible benchmark and extension for the research community.
\end{itemize}

This paper covers the background on fraud detection, including NLP/ML and multimodal solutions along with systematic reviews and cross-domain insights, in \autoref{sec:background}. In \autoref{sec:proposedSolution} we present the end-to-end pipeline for fraud detection using synthetic data. In \autoref{sec:experimental} we describe the experimental setup. \autoref{sec:results} discuss the results and outcome of the proposed solution followed by conclusion in \autoref{sec:conclusion}.

\section{Background and Related Work} \label{sec:background}
Multimodal data, AI, and NLP enables \emph{cross-modal verification}, linking voice identity with narrative similarity for fraud detection, but their public availability remains scarce, particularly in the insurance domain \cite{piehl2021classification,backlund2023detection}.
This scarcity is a recurring limitation in the literature and motivates synthetic multimodal datasets that capture fraud-related phenomena while respecting privacy constraints.

\begin{figure*}[!t]
\begin{center}
\includegraphics[width=1.65\columnwidth\relax]{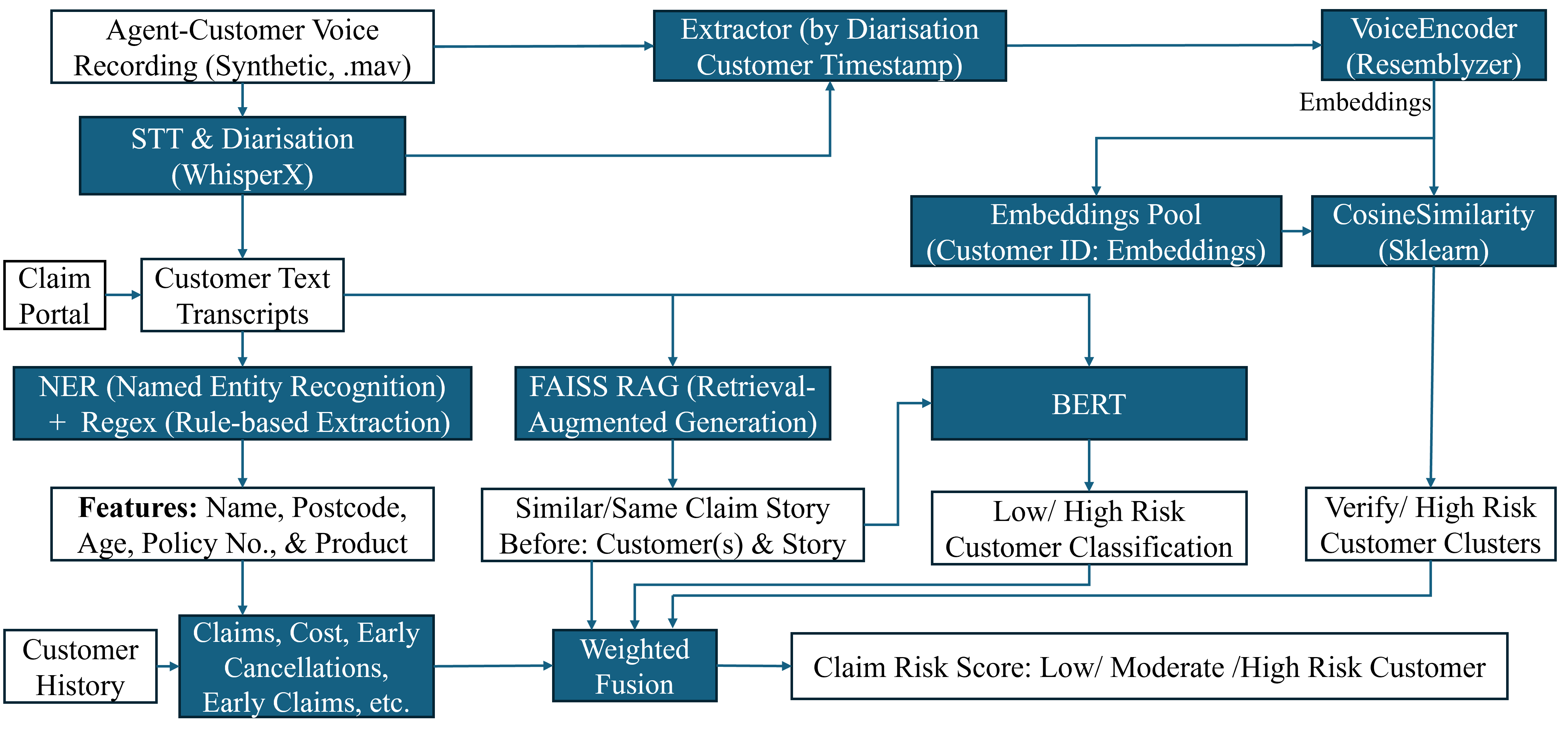}
\caption{Proposed end-to-end solution for AI-based robust insurance fraud claims detection.}
\label{fig:End2End}
\end{center}
\end{figure*}

\subsection{Insurance Fraud Detection Using NLP/ML:}
Pre-trained transformers like BERT \cite{backlund2023detection} and DistilBERT \cite{chang2022design} provide significant advantages due to their ability to capture context, bidirectional language understanding, transfer learning capabilities, ease of use, open-source availability, etc. These models can identify fraudulent indicators, inconsistencies, and patterns in customer conversations more effectively than traditional keyword-based approaches.

Early work on textual claim analysis applied bag-of-words and statistical classifiers to structured claim data, but transformers have since become prevalent. Piehl \cite{piehl2021classification} demonstrated that BERT-based models outperform traditional baselines for classifying Swedish insurance call transcripts into claim types, noting the detrimental effect of ASR noise on classification. B\"acklund and \"Ohman \cite{backlund2023detection} evaluated BERT, Word2Vec, and TF-IDF for fraud detection in textual and transcribed call data, reporting modest macro-F1 scores and emphasising the challenges of small, unbalanced datasets.

Recent research extends beyond classification to fraud awareness and explainability. Chang et al. \cite{chang2023design} proposed a chatbot to detect and classify financial fraud using NLP intent detection, while Dimri et al. \cite{dimri2024enhancing} explored domain-specific language models to support claim handlers in real-time. Gangani \cite{gangani2023ai} and Tarra \cite{tarra2024ai} highlight AI’s role in both detection and risk scoring, incorporating entity recognition and anomaly detection. Perumal \cite{perumal2023innovative} reviews innovative insurance-specific applications of AI, and Banulescu‐Radu and Yankol‐Schalck \cite{banulescu2023practical} provide operational guidelines for implementing ML-based fraud detection in household insurance.

\subsection{Fraud Detection in Call Data:}
Speech-based fraud detection addresses both insurance-specific and broader financial services scenarios. Leal et al. \cite{leal2018telephone} examined deception cues in insurance telephone calls, finding that structured ``model statements'' can elicit measurable linguistic differences. 
Kumar et al. \cite{kumar2022detecting} and Gupta \cite{gupta2024detection} addressed scam/spam call detection with NLP classifiers over ASR transcripts, a domain whose technical tools (ASR, NER, classification) transfer to insurance fraud use cases.

\subsection{\HL{Multimodal Fraud Solutions:}}
\HL{ 
In financial and call-centre contexts, acoustic and linguistic analyses have been applied to detect deception, although details are often proprietary }\cite{bajaj2020fraud}. \HL{More recent studies combine LLMs with interpretable machine learning to extract semantic and sentiment features from textual data, while noting limitations in interpretability, fraud-type granularity, and comparative validation} \cite{nie2025multimodal}. \HL{Healthcare fraud research likewise shows that combining structured claim data with narrative text, supported by preprocessing, feature engineering, and hybrid learning pipelines, improves anomaly detection} \cite{dupreez2023healthcare}. \HL{Within insurance and auto insurance, multimodal systems that fuse images, text, and tabular features achieve gains over unimodal baselines, but remain case-specific, rely on closed datasets and availability of all features, and do not include conversational call audio or aligned transcripts} \cite{yang2023auto,asgarian2023autofraudnet}. 

\HL{Commercial solutions also increasingly employ AI-powered fusion. Swiss Re’s \textit{ClaimsGenAI} automates claims processing from unstructured documents while maintaining human oversight, Xenoss integrates structured and unstructured streams such as transactions, behavioural signals, and text using semantic search and adaptive learning, and Moody’s \textit{Maxsight} unifies compliance, screening, and investigative workflows through explainable name matching \cite{swissre2025genai,xenoss2025fraud,moodys2025fraud}. However, these systems primarily operate at the document or transaction level and do not incorporate integrated speech–text processing at the FNOL stage. Collectively, they confirm the value of multimodal fusion but underscore the absence of reproducible, explainable, and shareable benchmarks that pair conversational FNOL audio with transcripts.}

\subsection{Systematic Reviews and Cross-Domain Insights:}
Systematic literature reviews in financial \cite{ali2022financial}, insurance \cite{aslam2022insurance}, e-commerce \cite{mutemi2023ecommerce}, and healthcare \cite{dupreez2023healthcare} fraud domains reveal following commonalities: (i) a lack of standardised, shareable datasets; (ii) challenges with class imbalance; (iii) the potential of hybrid models combining statistical, ML, and rule-based components. These reviews reinforce the importance of entity extraction \cite{boulieris2024fraudnlp}, history-aware scoring, and the multimodal evidence integration.

\subsection{ML Pipelines and Open Implementations}
Several open-source projects illustrate applied ML workflows for insurance fraud detection, including supervised classifiers, anomaly detection, and feature engineering \cite{nirab2021insuranceclaim,prapra2023insurancefraud,manoj2023predictive,cheuk2023frauddetector}. While not domain-complete, these repositories exemplify reproducible approaches and baseline architectures. Public datasets such as the Mendeley Insurance Claim Fraud dataset \cite{mendeley2020insurance} remain text-only and lack conversational speech, leaving a gap for multimodal synthetic dataset.

\section{Proposed End-to-End Pipeline} \label{sec:proposedSolution}
As outlined, existing works on insurance fraud detection is hindered by the absence of publicly shareable multimodal datasets, constrained by privacy and regulatory requirements. This limits progress even as fraudulent claims impose substantial financial losses and drive higher premiums. To address this gap, we present (\autoref{fig:End2End}) a synthetic, multimodal pipeline that emulates real FNOL conditions and supports reproducible research without exposing sensitive customer data.

Compared with prior studies, our approach unifies several components rarely combined in a single workflow: (i) generating \emph{synthetic} agent–customer dialogues containing structured entities, (ii) synthesising \emph{two-speaker audio} with xTTS, (iii) applying \emph{ASR + diarisation} via WhisperX to create realistic transcription conditions, (iv) integrating \emph{NER} and \emph{Regex} for feature extraction to fetch customer history, (v) integrating \emph{BERT}, \emph{RAG-based retrieval} and \emph{rule-based scoring} for classification, (vi) extracting \emph{speaker embeddings} for cross-case voice re-use detection, and (vii) fusing all of them into \emph{rule-based scoring} for risk scoring. This design explicitly addresses the need to minimise false positives, balancing fraud sensitivity with precision to safeguard customer experience at the critical claims stage.
To our knowledge, no previous work integrates these elements into a single end-to-end pipeline for insurance fraud detection.

\subsection{Synthetic Dataset Generation}
\begin{figure}[!t]
\begin{center}
\includegraphics[width=0.95\columnwidth]{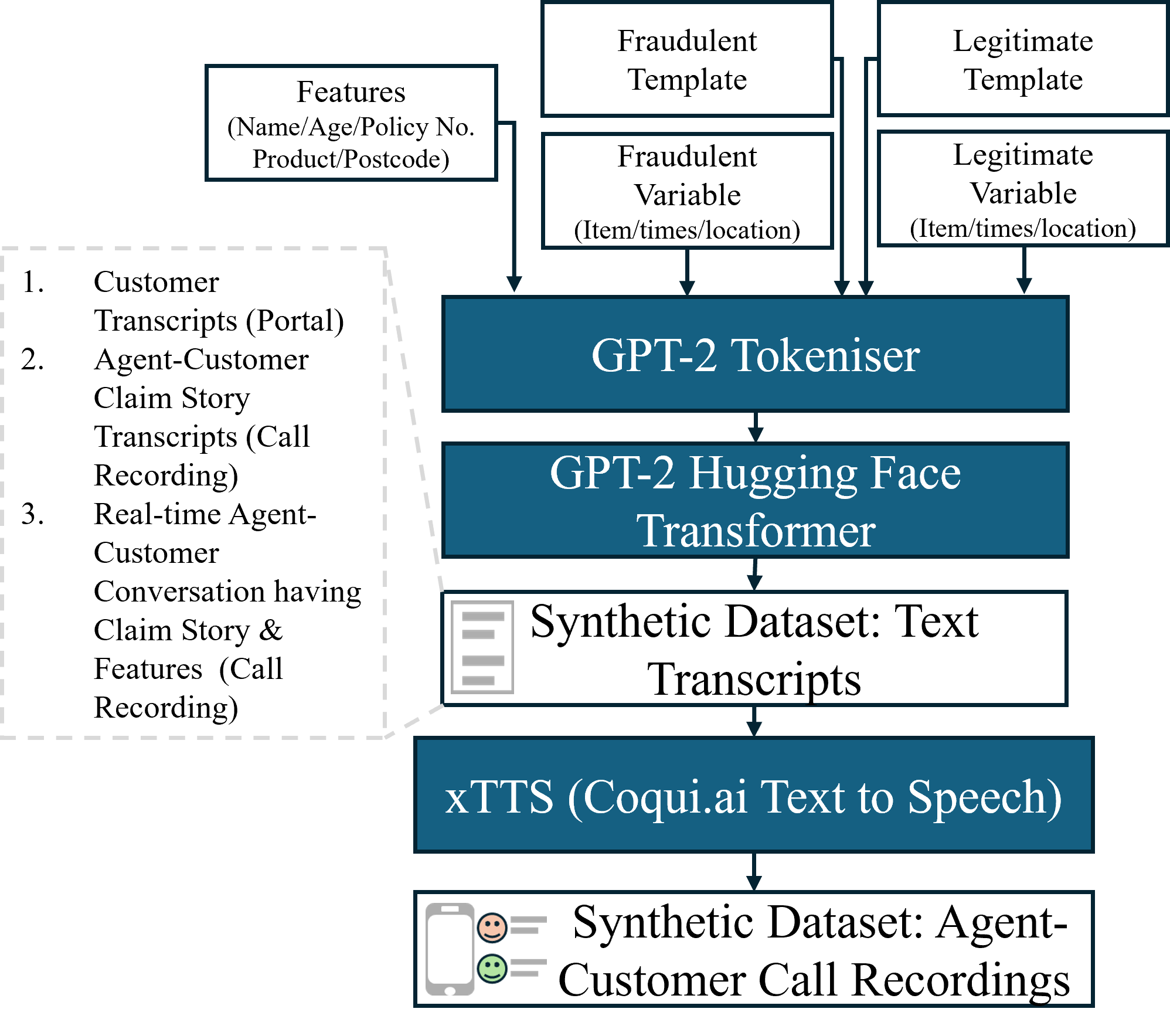}
\caption{Block diagram illustrating the process used for synthetic data generation.}
\label{fig:DataGeneration}
\end{center}
\end{figure}

\autoref{fig:DataGeneration} illustrates the synthetic data generation process. The pipeline begins with user-specified inputs, including (i) \emph{fraudulent and legitimate dialogue templates}, and (ii) \emph{structured variables} such as customer name, age, postcode, policy number, and product type. These inputs are passed to a GPT-2 transformer model, which generates diverse and coherent synthetic transcripts. 
Real FNOL calls often include small speech disfluencies like “um,” “uh,” restarts, and short pauses, which are typical in spontaneous conversation.
Our stochastic decoding (temperature sampling and nucleus/top‑k methods), encourages more varied wording and helps avoid rigid, repetitive patterns. As a result, the output sound more conversational and less formal.
The resulting transcripts are designed to emulate the types of narratives encountered in practice: (a) short customer-submitted text summaries via online portals; (b) agent–customer conversations recounting the details of a claim (when/where/what); and (c) full-length call centre dialogues that include both claim descriptions and associated~metadata.

Once generated, the transcripts are converted into audio using multi-speaker TTS synthesis (xTTS). This produces realistic two-speaker call recordings, with distinct agent and customer voices. The synthetic audio complements the text transcripts and replicates real-world FNOL channels, ensuring that the dataset reflects both written and spoken modalities of claim reporting. This multimodal design enables subsequent processing steps such as ASR, diarisation, entity extraction, and speaker embedding analysis.

\subsection{Transcription and Diarisation}
When claims are reported via telephone rather than through the portal, the first step (\autoref{fig:End2End}) is to convert audio into text and to extract customer speech. For this, we use WhisperX, which combines state-of-the-art ASR with alignment and diarisation. The model generates time-stamped transcripts and assigns speaker segments, while preserving conversational structure and introducing realistic transcription errors that mirror real-world call centre conditions. Diarisation is particularly important in the insurance setting, where claims are typically captured through two-party conversations, and downstream analysis often requires separating customer and agent contributions.

To identify the customer, diarisation output is first inspected to verify that multiple speakers were detected. By default, we assume that the agent speaks first and therefore label the second speaker as the customer. If this assumption fails, fallback heuristics can be applied: call-centre agents are typically limited in number and use scripted openings or identifiable key phrases, which allow them to be distinguished from customers. 
Using WhisperX word-level timestamps, we then extract the customer’s audio segments. These are passed to subsequent voice processing modules, including speaker embedding, similarity search, and clustering. This ensures that downstream fraud detection focuses on the customer’s speech (primary source), while agent speech serves mainly as context.

\vspace{4mm}
\subsection{Entity Extraction from Transcripts}
Following customer call transcription and recording, the textual channel is processed with NER combined with rule-based regular expressions to extract structured attributes such as \emph{name, postcode, age, product type, policy number, etc.}. 
This hybrid approach ensures robustness: NER provides generalisation across unseen contexts, while regex guarantees high precision for formalised identifiers like policy numbers or postal codes. 

Extracted entities are later used to query historical customer records and enrich the fraud scoring process with behavioral features such as prior claim counts, cancellations, missed payments, etc.

\subsection{Hybrid NLP-based Classification}
To analyse textual claim transcripts, we adopt a hybrid approach that combines transformer-based modelling with retrieval. BERT is fine-tuned for binary classification, using 2e-5 learning rate, batch size of 4, and 0.01 weight decay to mitigate overfitting. Prior to training, all transcripts are tokenised using the BERT tokeniser to convert text into numerical representations \cite{deep_learning_nlp_guide}.

To capture narrative repetition across customers, we integrate RAG via LangChain. Customer transcripts are embedded and indexed with FAISS (Facebook AI Similarity Search), enabling efficient similarity search. At inference, the model retrieves semantically related past claims, which improves contextual grounding and supports the detection of organised fraud patterns, such as groups of customers reusing the same story until it fails \cite{ali2022financial}. This combination allows the system to leverage both pre-trained knowledge and case-specific history. Confidence scoring is applied to ensure that only high-certainty predictions are flagged, presenting a systematic approach.

\subsection{Speaker Clustering}
Beyond lexical content, the customer’s voice provides an additional modality for fraud detection. To capture speaker identity, we use Resemblyzer, which encodes utterances into fixed-dimensional embeddings. Cosine similarity between embeddings enables detection of repeated voices across different claims, even when associated with different customer details. This allows identification of high-risk patterns, such as organised groups recycling the same narratives or a single speaker impersonating multiple customers. Incorporating speaker identity complements text-based analysis and supports cross-modal fraud evidence.

\subsection{Fraud Risk Scoring Framework}
\begin{figure}[!t]
\begin{center}
\includegraphics[width=0.9\columnwidth]{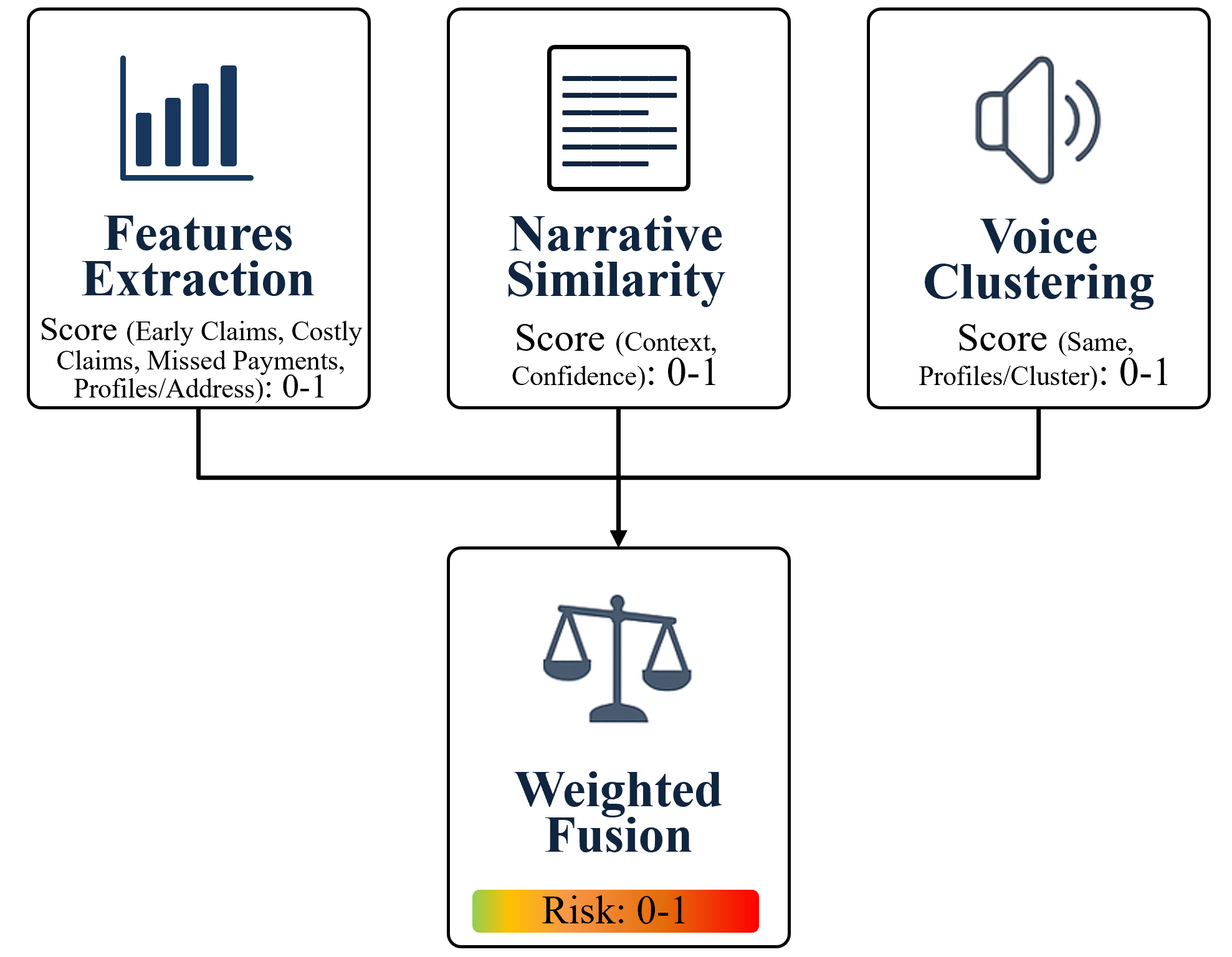}
\caption{Block diagram illustrating the fraud risk scoring framework.}
\label{fig:RiskFramework}
\end{center}
\end{figure}

The final stage of the pipeline fuses outputs from text, audio, and structured features into a single interpretable fraud risk score. Given the high-stakes nature of fraud detection, where both prevention and customer satisfaction are critical, the framework incorporates additional reliability mechanisms (\autoref{fig:RiskFramework}). Rather than relying on a single model, we integrate four complementary signals: (i) structured feature checks, (ii) semantic narrative similarity, (iii) voice similarity and clustering, and (iv) confidence-weighted fusion. Each component is normalised onto a $[0,1]$ scale, after which a weighted aggregation produces the final score. This design ensures that different aspects of fraud behaviour (claims history, narrative reuse, speaker identity) are jointly evaluated while remaining transparent to investigators.

\subsubsection{Structured feature extraction.}  
Customer histories are retrieved using extracted entities (policy number, name, postcode). From these records, the following risk indicators are computed:
\begin{enumerate}
    \item \textbf{Early claims:} whether a claim is filed within 30 days of policy inception.  
    \item \textbf{High-cost claims:} whether the claim amount exceeds $1.2 \times$ the average claim cost for the same product type.  
    \item \textbf{Payment irregularities:} number of missed payments or direct debit rejections $> 5$.  
    \item \textbf{Household profile inflation:} number of different customer profiles/postcode $> 4$ (assuming  distinct postcode/household). The threshold of 4 is motivated by the UK Office for National Statistics report (2024), which shows over 93.6\% of households have 1-4 members \cite{ONS2025Families}.
\end{enumerate}
Each sub-condition is scored between 0 (no anomaly) and 1 (clear anomaly), allowing a composite \emph{feature-risk score} to be calculated as their average.

\subsubsection{Semantic narrative similarity (Hybrid NLP).}  
The BERT-RAG module retrieves similar transcripts from the historical database indexed with FAISS. A claim is flagged as suspicious if a retrieved transcript has $>90\%$ similarity (context) and the model’s confidence score exceeds 0.9. The resulting \emph{textual-risk score} is normalised to [0,1] based on similarity magnitude and confidence level, with higher values representing stronger evidence of narrative reuse across customers.

\subsubsection{Voice similarity and clustering.}  
Speaker embeddings (Resemblyzer) are used to identify repeated voices across claims. Two conditions are applied: (i) cosine similarity $>0.75$ between embeddings, and (ii) the customer belongs to a cluster of more than four profiles. The threshold of 4 again reflects the maximum household size assumption. If both conditions are satisfied, the \emph{voice-risk score} is scaled toward 1, with lower similarity or smaller cluster sizes producing proportionally reduced scores.

\subsubsection{Weighted risk fusion.}  
The final fraud risk score is obtained as a weighted combination of the three modalities:
\begin{equation}
Risk = w_f \cdot S_{\text{features}} + w_s \cdot S_{\text{similarity}} + w_v \cdot S_{\text{voice}}
\end{equation}
where $w_f$, $w_s$, and $w_v$ are weights for features, text similarity, and voice respectively, with $w_f + w_t + w_v = 1$. In practice, higher weight is assigned to structured features (e.g., $w_f=0.4$) since they capture objective financial behaviour, while text and voice cues are complementary ($w_t=w_v=0.3$ each). The score is then mapped to risk bands (low, medium, high) for operational interpretability.

This weighted, multi-signal approach avoids over-reliance on any single modality and provides investigators with an interpretable audit trail: a flag can be traced back to its textual similarity, anomalous features, or repeated speaker identity.

\section{Experimental Setup}\label{sec:experimental}
The pipeline is implemented entirely with open-source frameworks to ensure reproducibility. Core components from Hugging Face Transformers \cite{wolf-etal-2020-transformers} provide pre-trained BERT \cite{devlin2019bert} and GPT-2 models \cite{radford2019language, liu2024gpt}, reducing training cost and enabling domain adaptation for insurance narratives. GPT-2 (GPT2LMHeadModel), paired with its tokeniser, generates balanced legitimate and fraudulent dialogues with realistic lexical~variety.

For speech synthesis, Coqui’s xTTS \cite{coqui2023xtts, casanova2024xtts} creates natural two-speaker audio to emulate agent–customer calls, while Google’s gTTS \cite{gtts2024} is used for single-speaker feature-only utterances where rapid generation suffices. Synthetic recordings are transcribed and diarised with WhisperX \cite{bain2023whisperx}, chosen for its word-level alignment and reliable speaker attribution, both critical in call-centre scenarios.

Downstream text analysis combines RoBERTa-base NER \cite{roberta_large_ner_english} with regex rules to capture structured identifiers (e.g., policy numbers, postcodes) and support linkage to simulated customer histories. Sentence-Transformer embeddings (all-MiniLM-L6-v2 \cite{sentence-transformers-all-minilm-l6-v2}) are indexed in FAISS \cite{douze2024faiss,johnson2019billion} for fast semantic retrieval, enabling RAG \cite{lewis2020retrieval} via LangChain \cite{langchain_repo}. BERT is fine-tuned for binary fraud classification, leveraging contextual embeddings beyond keyword heuristics.

For the audio channel, Resemblyzer \cite{resemblyzer2020} produces fixed-length speaker embeddings; cosine similarity (scikit-learn \cite{scikit-learn}) scores detect potential voice reuse across claims and support clustering. All outputs, structured features, transcript similarity, and speaker scores, feed a rule-based fusion layer for interpretable risk scoring. This configuration balances fidelity to FNOL conditions with efficiency, while remaining easily extensible for future experiments.

\section{Results and Discussion}\label{sec:results}
We present the results associated with each of the components in the end-to-end pipeline, demonstrating the system architecture's plausibility. 

\subsection{Generated Synthetic Datasets}\label{sec:generateddata}
The generated dataset is designed to balance fraudulent and legitimate claims while reflecting the multimodal nature of FNOL data. In total, it comprises:

\begin{enumerate}
    \item \textbf{500 customer-only transcripts} (balanced fraudulent and legitimate) are intended for training and evaluating text-based models such as BERT-RAG. Their primary use case is retrieving similar narratives across customers, which can help uncover repeated stories used by organised fraud groups and to fine-tune BERT like models for binary classification. 
    
    \item \textbf{250 agent–customer transcripts} (balanced fraudulent and legitimate) with corresponding \textbf{250 recordings generated using TTS} having 12 distinct speakers. 
    This subset supports research on speech processing, transcription, and diarisation. By simulating conversational exchanges, it allows the evaluation of ASR performance and the accuracy of the customer-agent separation.
    
    \item \textbf{9 feature-only customer texts} paired with \textbf{9 recordings using gTTS} (1 speaker). 
    These samples focus on structured attribute extraction (name, age, postcode, product, policy number) and provide a baseline for assessing transcription robustness and NER precision in minimal-text settings.
    
    \item \textbf{250 feature-rich agent–customer transcripts} (balanced fraudulent and legitimate) with \textbf{250 recordings generated using xTTS} across 38 speakers. 
    This subset represents the most complete simulation of FNOL calls, combining conversational narratives with embedded features. It is designed for joint evaluation of transcription, diarisation, entity extraction, clustering, and fraud scoring.
\end{enumerate}
Together, these subsets support multimodal benchmarking under realistic FNOL conditions.

\begin{table*}[!t]
\centering 
\caption{\label{tab:feature} Results across all processing stages, from voice recording to extracted features. The table reports both intrinsic dataset validation metrics and model performance measures. Here, WER: Word Error Rate, N: Name, Pe: Postcode, A: Age, Py: Policy No, Pt: Product, Std: Standard Deviation, TPU: Tokens/Utterance, TPD: Turns/Dialogue, TTR: Type-Token Ratio (Lexical diversity), and AD: Audio~Duration.}
\vspace{1mm}
\resizebox{\textwidth}{!}{%
\begin{tabular}{@{\hspace{0em}}c@{\hspace{0.4em}}c@{\hspace{0.4em}}c@{\hspace{0.4em}}c@{\hspace{0.4em}}c@{\hspace{0.4em}}c@{\hspace{0.4em}}c@{\hspace{0.4em}}c@{\hspace{0.4em}}c@{\hspace{0.4em}}c@{\hspace{0em}}}
\hline
\textbf{} & \textbf{Voice$\rightarrow$Transcribed} & \textbf{Text$\rightarrow$Features} &\textbf{Voice$\rightarrow$Features}&  \multicolumn{3}{c}{\textbf{\HL{Voice$\rightarrow$Features}}} & \multicolumn{3}{c}{\textbf{Voice$\rightarrow$Features}}\\
\hline
Test Samples &   \cite{commonvoice} 136& (GPT2-based) 9&(gTTS-based) 9&   \multicolumn{3}{c}{\HL{(TTS-based) 250}} & \multicolumn{3}{c}{(xTTS-based) 250
}\\
\makecell{Speakers/Samples} &  1& -&1&   \multicolumn{3}{c}{\HL{2}} & \multicolumn{3}{c}{2
}\\
\makecell{Overall Speakers} &  39& -&1&   \multicolumn{3}{c}{\HL{12}} & \multicolumn{3}{c}{38
}\\
\HL{Uniqueness N/Py}&\HL{-}&\HL{100/89\%}
&\HL{100/89\%}&\multicolumn{3}{c}{\HL{0.4/-\%}}&\multicolumn{3}{c}{\HL{12/91\%}}\\
\HL{Valid Pe/Py}&\HL{-}&\HL{89/89\%}
&\HL{89/89\%}&\multicolumn{3}{c}{\HL{-/-}}&\multicolumn{3}{c}{\HL{87/91\%}}\\
\HL{Entropy Pt}&\HL{-}&\HL{1.83}
&\HL{1.83}&\multicolumn{3}{c}{\HL{2.83}}&\multicolumn{3}{c}{\HL{3.03}}\\
\HL{Mean±Std A}&\HL{-}&\HL{40±10}&\HL{40±10}&\multicolumn{3}{c}{\HL{-±-}}&\multicolumn{3}{c}{\HL{47±18}}\\
\HL{Mean±Std TPU/TPD/TTR}&\HL{-}&\HL{-}&\HL{-}&\multicolumn{3}{c}{\HL{6.15±0.82/8.54±1.5/0.76±0.04}}&\multicolumn{3}{c}{\HL{6.56±0.65/8.54±1.5/0.69±0.03}}\\
\HL{Mean±Std AD}&\HL{-}&\HL{-}&\HL{-}&\multicolumn{3}{c}{\HL{55.89±11.09 sec}}&\multicolumn{3}{c}{\HL{24.12±3.42 sec}}\\
Accuracy &  99.99\%& 100\%&99.99\%&\makecell{\HL{93.24\%}\\\HL{Transcribed}}&\makecell{\HL{46.62\%}\\\HL{Diarised}}&\makecell{\HL{93.24\%}\\\HL{Features}}& \makecell{99.2\%\\Transcribed}& \makecell{88.4\%\\Diarised}&\makecell{90.4\%\\Features} \\
\HL{STT WER}&\HL{11.9\%}&\hl({-}&\HL{15.9\%}&\multicolumn{3}{c}{\HL{6.6\%}}&\multicolumn{3}{c}{\HL{20.4\%}}\\
\HL{Precision N/Pe/A/Py/Pt}&\HL{-}&\HL{100\%}&\HL{100\%}&\multicolumn{3}{c}{\HL{100\%}
}&\multicolumn{3}{c}{\HL{100\%}}\\
\HL{Recall N/Pe/A/Py/Pt}&\HL{-}&\HL{100\%}&\HL{77.8/100/100/88.9/100\%}&\multicolumn{3}{c}{\HL{96.6/-/-/-/63.8\%}}&\multicolumn{3}{c}{\HL{74.5/31.2/74.9/21.9/85.4\%}}\\
\HL{F1 N/Pe/A/Py/Pt}&\HL{-}&\HL{100\%}&\HL{87.5/100/100/94.1/100\%}&\multicolumn{3}{c}{\HL{98.3/-/-/-/77.9\%}}&\multicolumn{3}{c}{\HL{85.4/47.5/85.9/35.9/92.1\%}}\\
\hline
\end{tabular}} 
\label{tab:Evaluation}
\end{table*}

\HL{Since no equivalent public multimodal dataset is available, we evaluate the generated synthetic datasets using the metrics summarised in} \autoref{tab:Evaluation}. \HL{These include the number of test samples, speakers per sample, total unique speakers, uniqueness of names and policy numbers, validity of postcodes and policy numbers, product entropy, mean and standard deviation of age, tokens per utterance, turns per dialogue, lexical diversity, and audio duration. The evaluation reflects that the synthetic data is internally consistent, operationally realistic, and representative of plausible FNOL scenarios, while safeguarding sensitive information.}

\subsection{Feature Extraction}
Before diving into the feature extraction, the implementation was tested for transcription, text to feature extraction, voice transcription for feature extraction from the voice, then voice to transcription with diarisation for feature extraction. 
The process implementation chunks were tested on different dataset, starting from 136 samples of common voice dataset (\cite{commonvoice} Delta Segment 19.0) for transcription
, as the voices from common voice dataset were used as reference for customer-agent synthetic recordings. 

\begin{figure}[!t]
    \centering
    \vspace{2mm}
    \includegraphics[width=1\linewidth]{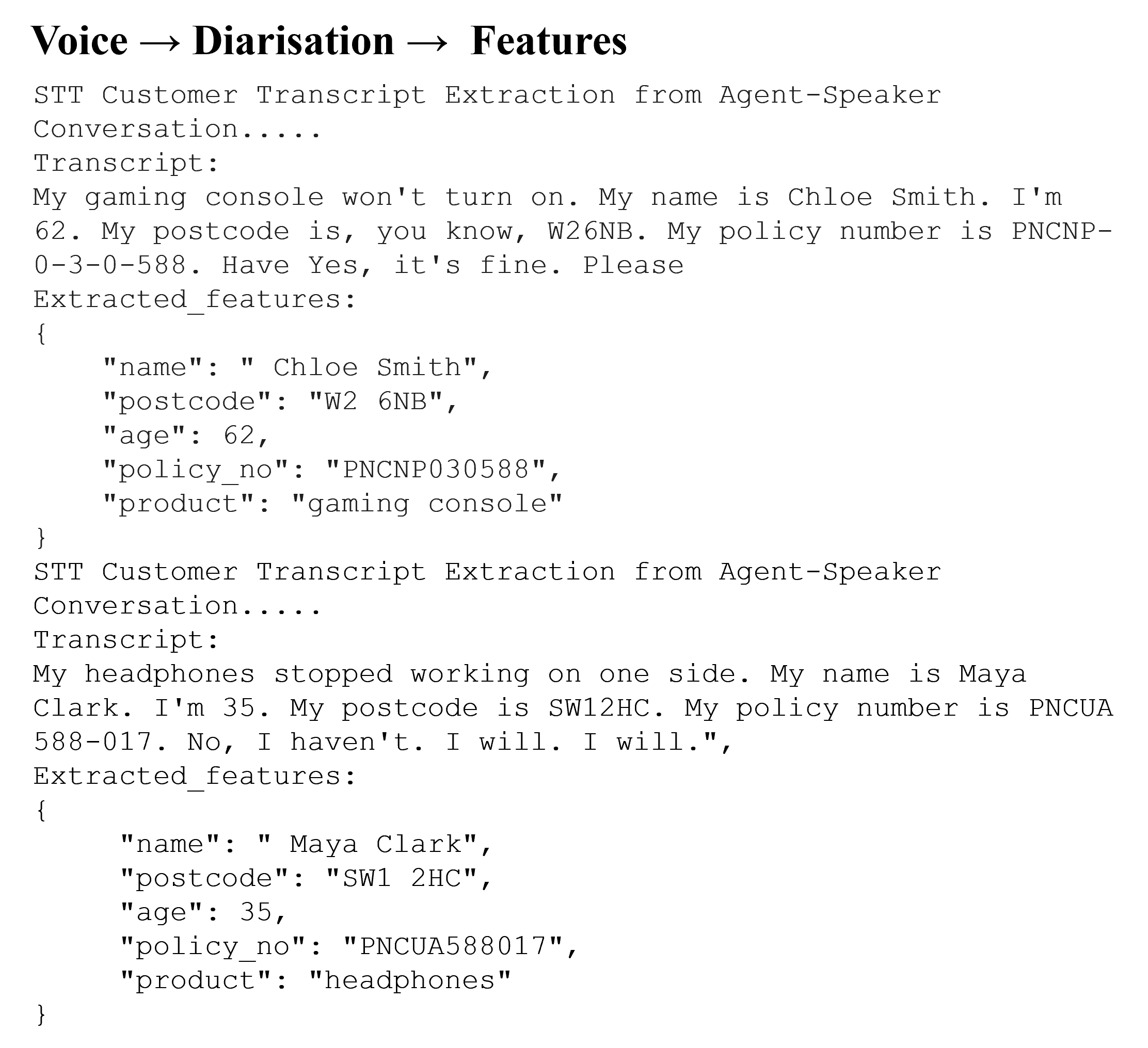}
    \caption{Feature extraction from xTTS based customer-agent recordings.}
    \label{fig:xTTsf}
\end{figure}

By using the process defined in \autoref{fig:End2End}, from synthetically generated 9 feature-only customer texts and its paired recording were tested for the feature extraction from text and recordings, 
highlights the feature extraction performance. 
The next evaluation, presents the complete evaluation of the feature extraction system. Where xTTS-based customer-agent recordings are used for the transcription, diarisation, and feature extraction. \autoref{fig:xTTsf} presents the highlights of the process and performance. 

\HL{Finally,} \autoref{tab:Evaluation} \HL{presents the overall performance in all stages of the processing, including text transcription, feature extraction from text and voice, and analysis of agent-customer recordings. The evaluation spans the entire pipeline, reporting the accuracy of transcription, speaker diarisation and feature extraction, the STT word error rate (WER) and precision, recall and F1 scores for the extracted features. 
It is important to note that the WER is computed using STT transcriptions of fully synthetic audio generated from synthetic text. The results under these conditions highlight the effectiveness of the proposed workflow and its potential for further improvement with even more enhanced synthesis quality.}

\begin{figure*}[!t]
    \centering
    \vspace{2mm}
    \subfloat{\includegraphics[width=0.33\linewidth]{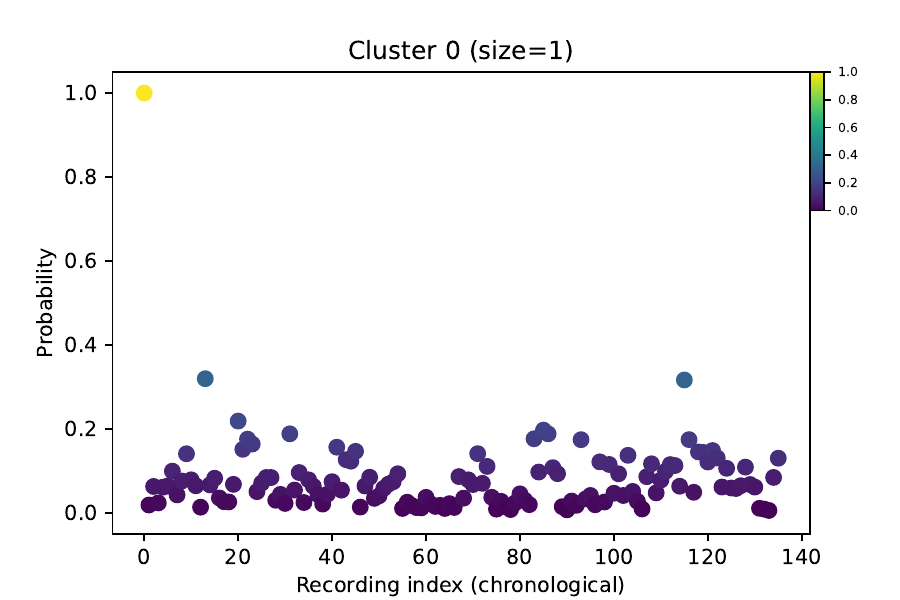}%
    \label{fig:c0}}\hfil
    \subfloat{\includegraphics[width=0.33\linewidth]{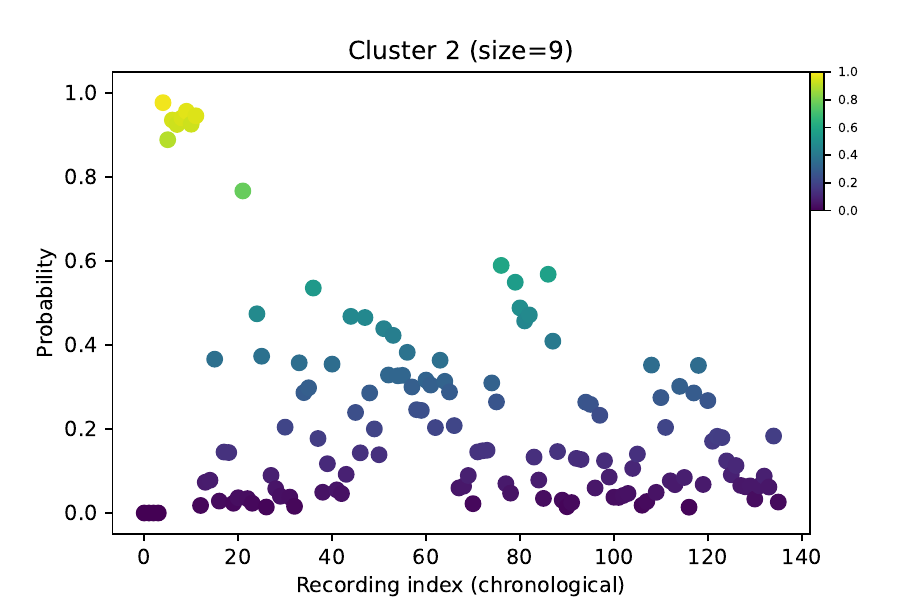}%
    \label{fig:c2}}\hfil
    \subfloat{\includegraphics[width=0.33\linewidth]{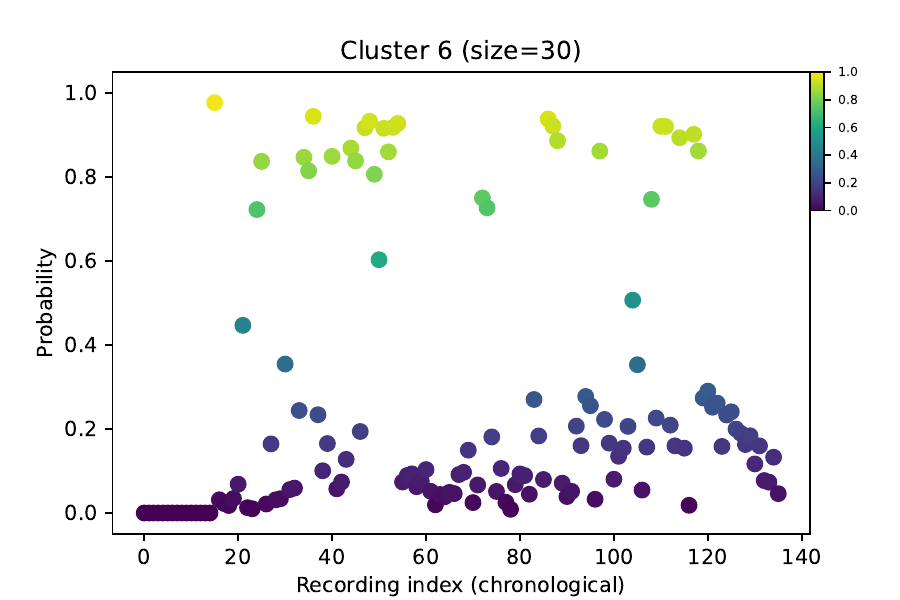}%
    \label{fig:c6}}
    \caption{Cluster formation on Common Voice samples (39 unique speakers, 136 samples).}
    \label{fig:c026}
\end{figure*}

\begin{figure}[!t]
    \centering \vspace{0mm}
    \includegraphics[width=\linewidth]{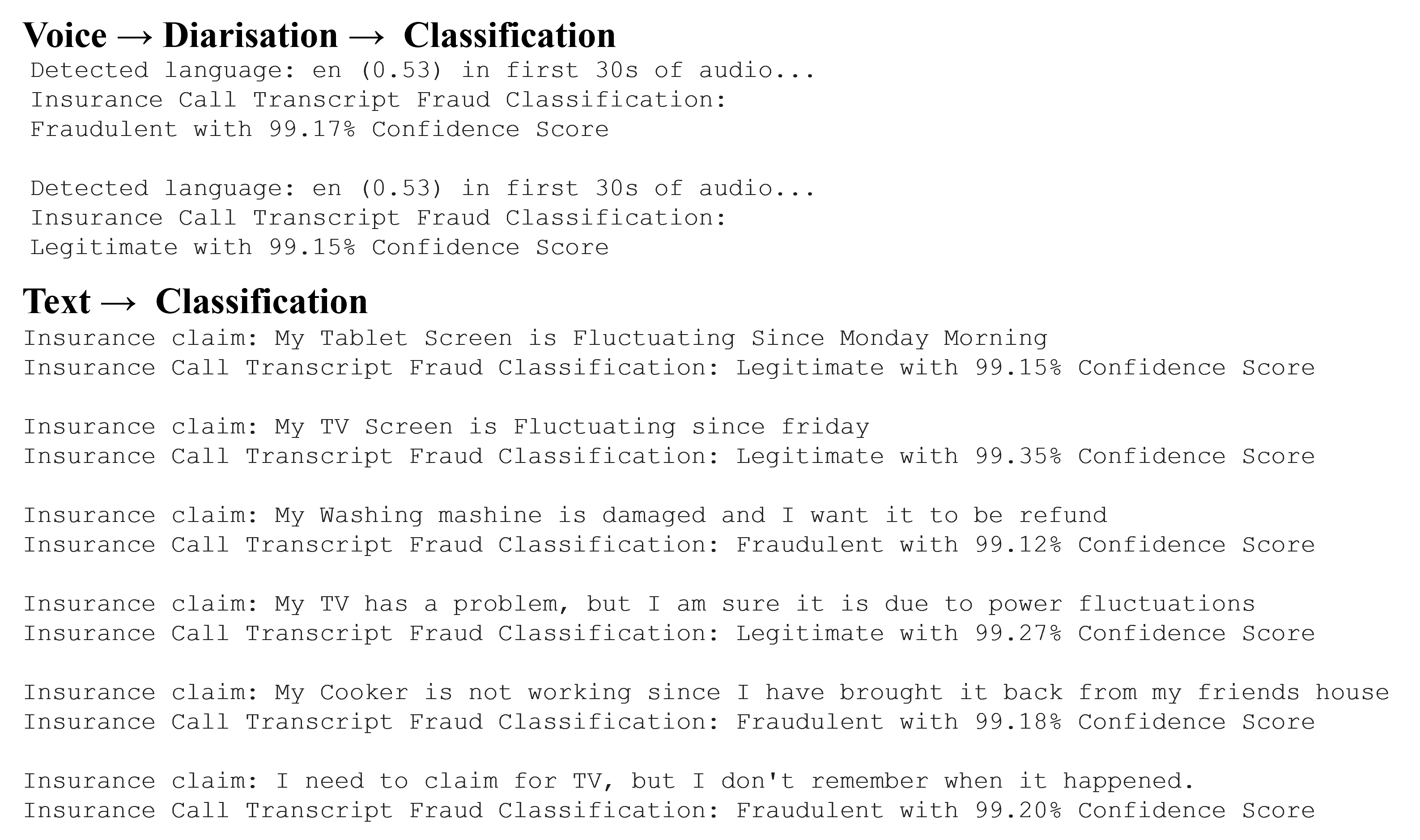}
    \vspace{1.5mm}
    \caption{Binary classification applied to claim-portal–style customer transcripts (GPT-2) and agent–customer call (xTTS) scenarios.}
    \vspace{3mm}
    \label{fig:UIdemo}
\end{figure}

\vspace{2mm}
\subsection{Binary Classification}
The BERT–RAG classifier was trained on the claim–portal–style customer-only transcripts comprising 500 synthetically generated samples (\autoref{sec:generateddata}), evenly balanced between fraudulent and legitimate classes. The dataset was partitioned using an 80/20 hold-out strategy, allocating 64\% for training, 16\% for validation, and 20\% for testing. During training, the validation loss progressively decreased from 0.646 to 0.0083, demonstrating stable convergence. In the synthetic hold-out test set, the model achieved a test loss of 0.0085 and near-perfect performance on all metrics: accuracy, precision, recall, F1 and AUC all at 100\%. However, these results should be interpreted with caution: The synthetic dataset is limited in size and controlled in variability, which yields optimistic performance estimates. 

\begin{table}
    \centering 
    \vspace{2mm}
    \caption{\HL{Performance evaluation of the binary classifier on a totally unseen dataset and format. Here, hyp: STT-generated transcripts,  ref: original reference transcripts, d: diarised, and u: undiarised.} 
    \vspace{1mm}
    }
    \resizebox{\columnwidth}{!}{
    \begin{tabular}{@{\hspace{0em}}c@{\hspace{0.3em}}c@{\hspace{0.3em}}c@{\hspace{0.3em}}c@{\hspace{0.3em}}c@{\hspace{0em}}c@{\hspace{0em}}}\hline
         & \makecell{\HL{Accuracy}\\(\%)} & \makecell{\HL{Precision}\\(\%)} & \makecell{\HL{Recall}\\(\%)} & \makecell{\HL{F1}\\(\%)} & \makecell{\HL{Avg. Confidence}\\ (\%)} \\\hline
         \HL{TTS($\text{hyp}_d$,$\text{ref}_d$)} & \HL{33,23} & \HL{20,25} & \HL{42,76} & \HL{28,37} & \HL{87,90} \\
         \HL{TTS($\text{hyp}_u$,$\text{ref}_u$)} & \HL{66,67} & \HL{62,63} & \HL{82,83} & \HL{71,72} & \HL{89,88} \\
         \HL{xTTS($\text{hyp}_d$,$\text{ref}_d$)} & \HL{53,51} & \HL{52,51} & \HL{95,95} & \HL{67,66} & \HL{93,93} \\
         \HL{xTTS($\text{hyp}_u$,$\text{ref}_u$)} & \HL{63,60} & \HL{58,57} & \HL{87,87} & \HL{67,69} & \HL{92,90} \\
         \HL{gTTS($\text{hyp}$,$\text{ref}$)} & \HL{89,89} & \HL{-}& \HL{-}& \HL{-}& \HL{96,95} \\\hline
    \end{tabular}
    }
    \label{tab:binary}
\end{table}

\HL{Following the synthetic proof-of-concept experiments, the classifier was evaluated on totally unseen datasets differing in both format and acoustic characteristics. As summarised in } \autoref{tab:binary}, \HL{these evaluations encompassed original (ref) and STT-based (hyp) transcripts under both diarised (customer-only, d) and undiarised (u) conditions, across three TTS variants, TTS, xTTS, and gTTS (which is already customer-only).
The results indicate that the classifier maintained coherent decision boundaries even when faced with introduced variations in contrast to the model training data. The same is suggested by the high recall values that model remains sensitive to fraud-related cues, though with reduced precision. }

We therefore present these results as a baseline and component-level proof-of-function,  requiring further work to ensure robustness on heterogeneous, real-world data. Improving generalisation will require (i) greater narrative and voice diversity in synthetic data, (ii) k-fold cross-validation, and (iii) fine-tuning on anonymised real claims when available.

\begin{figure}[!t]
    \centering \vspace{0mm}
    \includegraphics[width=1\linewidth]{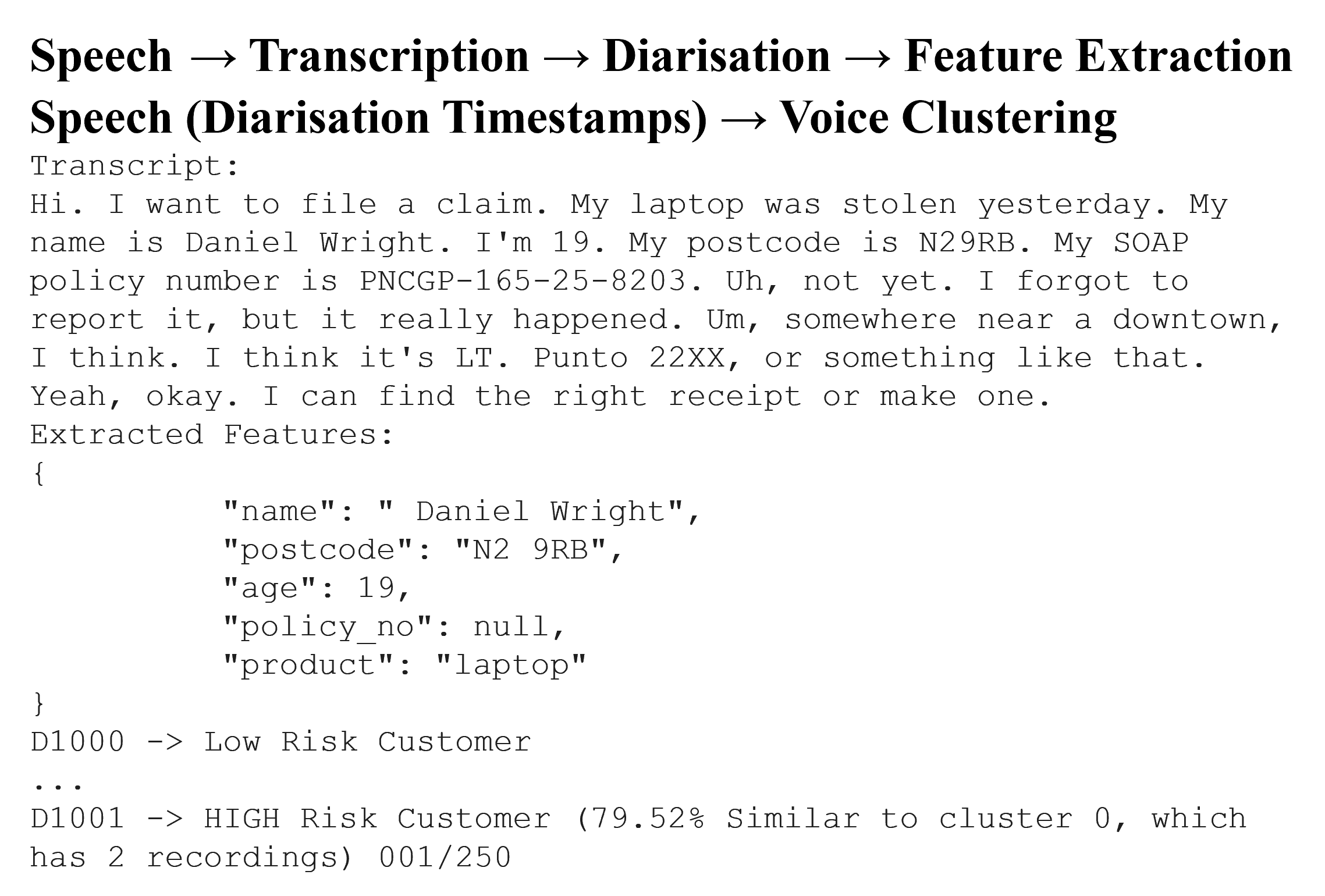}
    \caption{Voice clustering of xTTS-based agent–customer recordings.}
    \label{fig:voiceclustering}
\end{figure}

\autoref{fig:UIdemo} highlights the classification workflow across both claim portal-like textual submissions and call-centre-like dialogues, demonstrating seamless integration of transcription, diarisation, and fraud labelling.

\subsection{Voice Clustering}
Speaker embeddings derived from diarisation timestamps are used to isolate customer segments for verification and clustering. This enables detection of repeated speakers across claims, an indicator of potential organised fraud. \autoref{fig:voiceclustering} illustrates the workflow: ASR $\rightarrow$ diarisation $\rightarrow$ feature~extraction $\rightarrow$ similarity-based clustering.

\vspace{1mm}
Cosine similarity between embeddings is converted to a probability via a sigmoid mapping:
\begin{equation}
p(x) = \frac{1}{1 + e^{-k (x - t)}}
\end{equation}
where $x$ is the similarity score, $t$ the decision threshold (set at 0.75), and $k$ controls curve sharpness. Probabilities above 0.5 are grouped as the same speaker, supporting cluster formation across multiple cases.

\begin{figure*}
\centering
\subfloat{\includegraphics[width=0.25\linewidth]{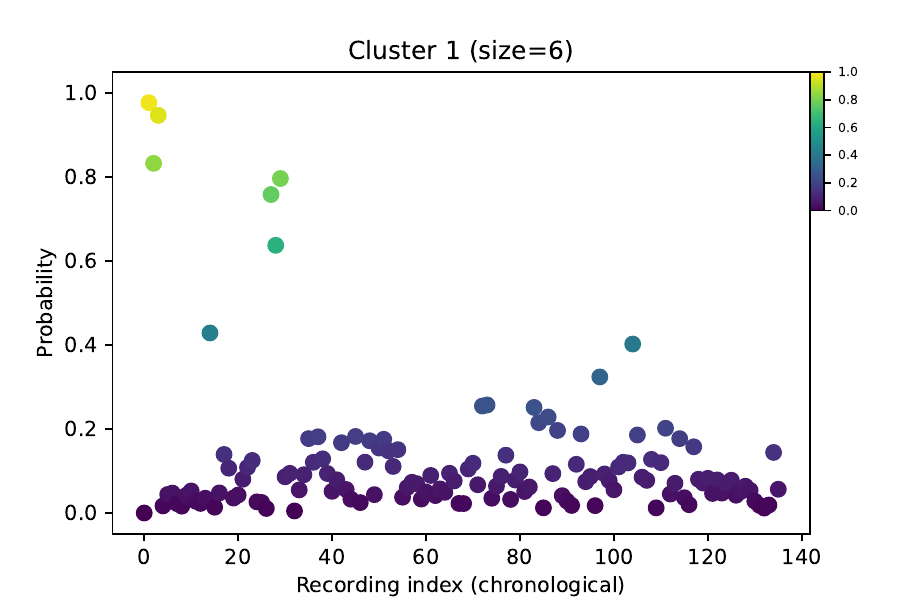}%
\label{fig:c1}}\hfil
\subfloat{\includegraphics[width=0.25\linewidth]{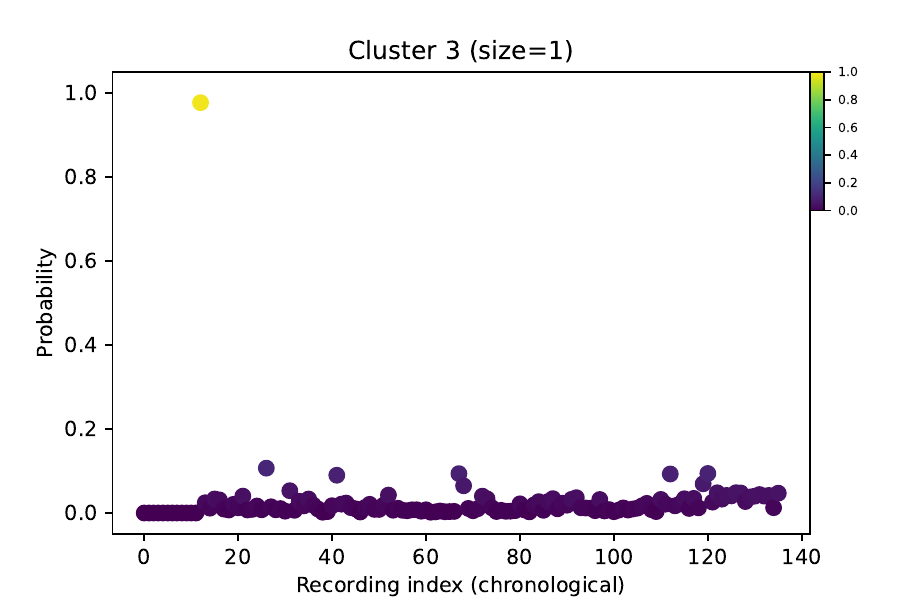}%
\label{fig:c3}}\hfil
\subfloat{\includegraphics[width=0.25\linewidth]{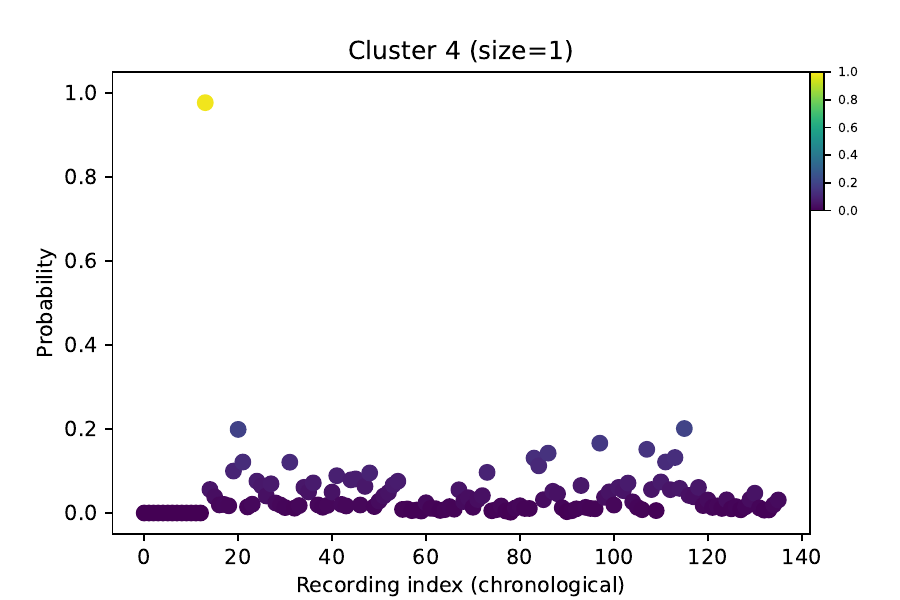}%
\label{fig:c4}}\hfil
\subfloat{\includegraphics[width=0.25\linewidth]{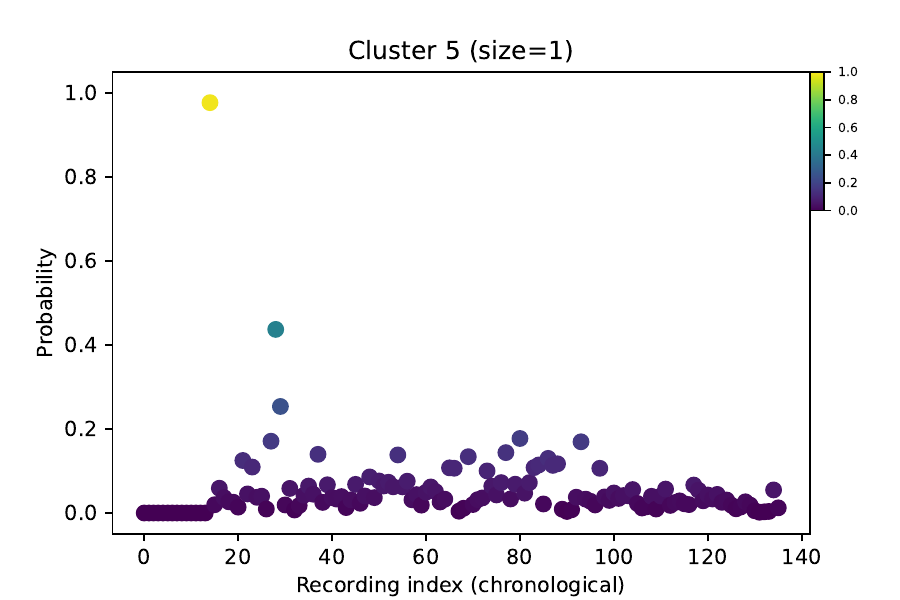}%
\label{fig:c5}}
\vspace{-1.85em}
\subfloat{\includegraphics[width=0.25\linewidth]{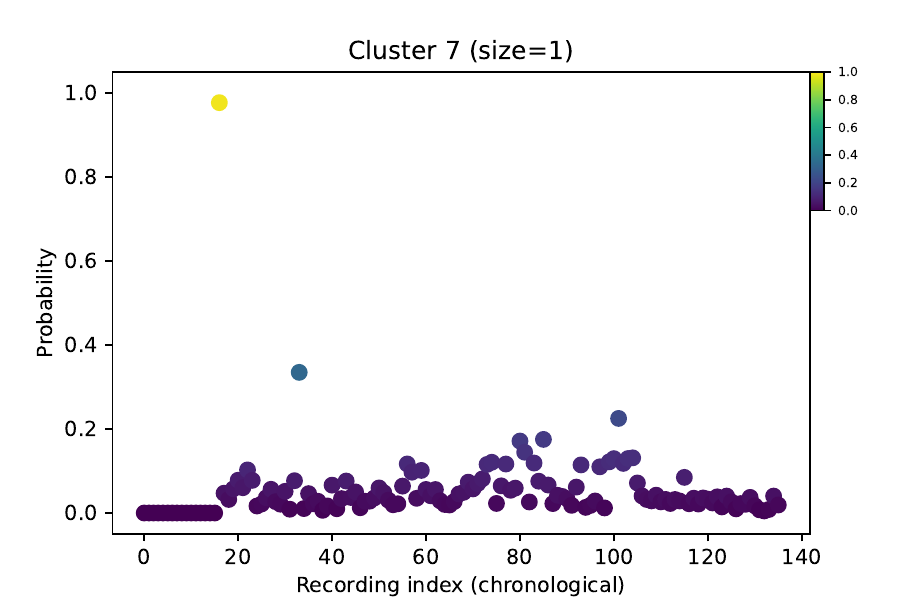}%
\label{fig:c7}} \hfil
\subfloat{\includegraphics[width=0.25\linewidth]{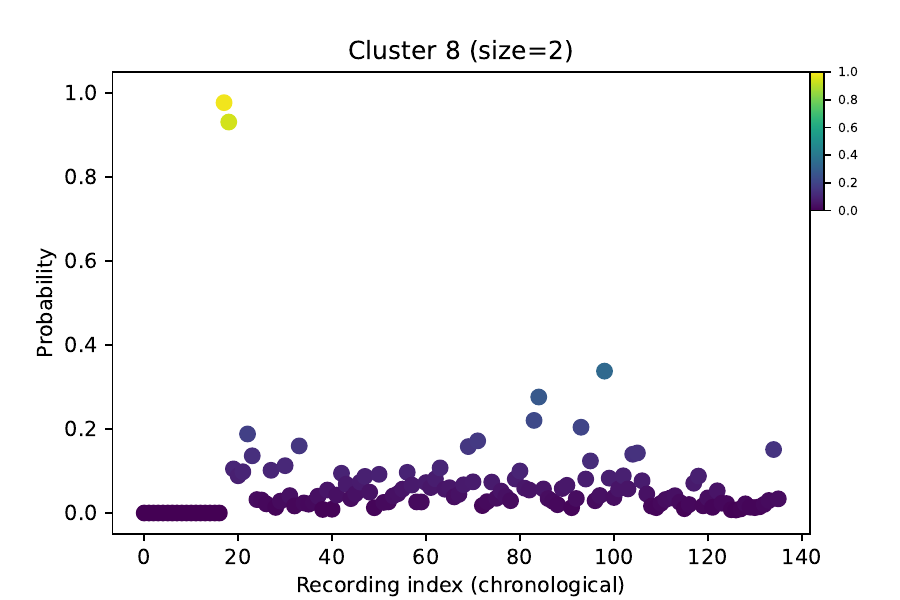}%
\label{fig:c8}}\hfil
\subfloat{\includegraphics[width=0.25\linewidth]{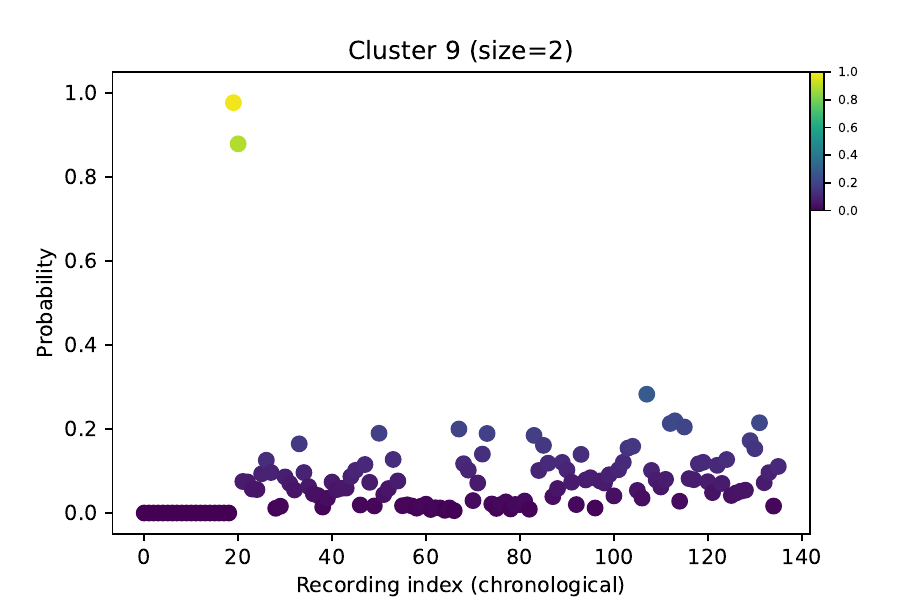}%
\label{fig:c9}}\hfil
\subfloat{\includegraphics[width=0.25\linewidth]{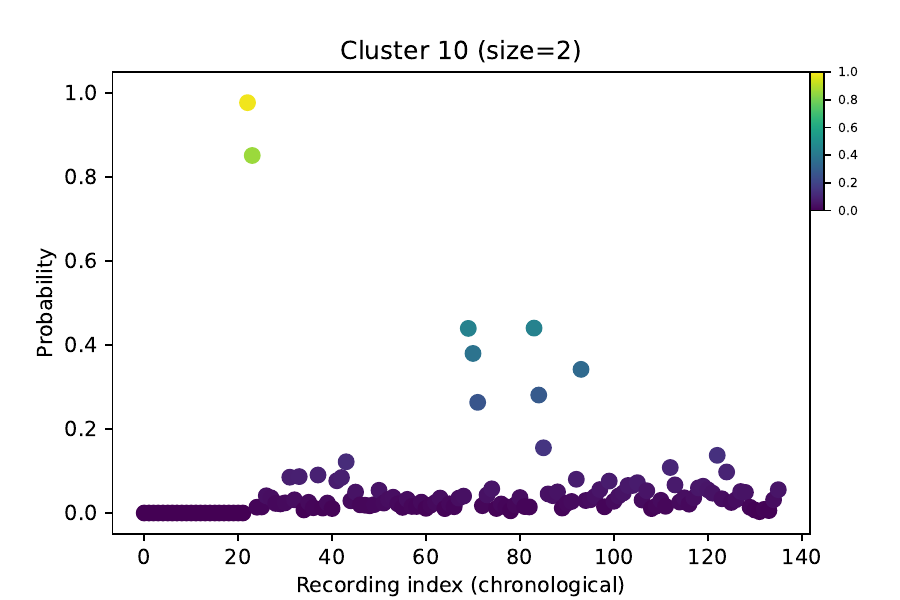}%
\label{fig:c10}}
\vspace{-1.85em}
\subfloat{\includegraphics[width=0.25\linewidth]{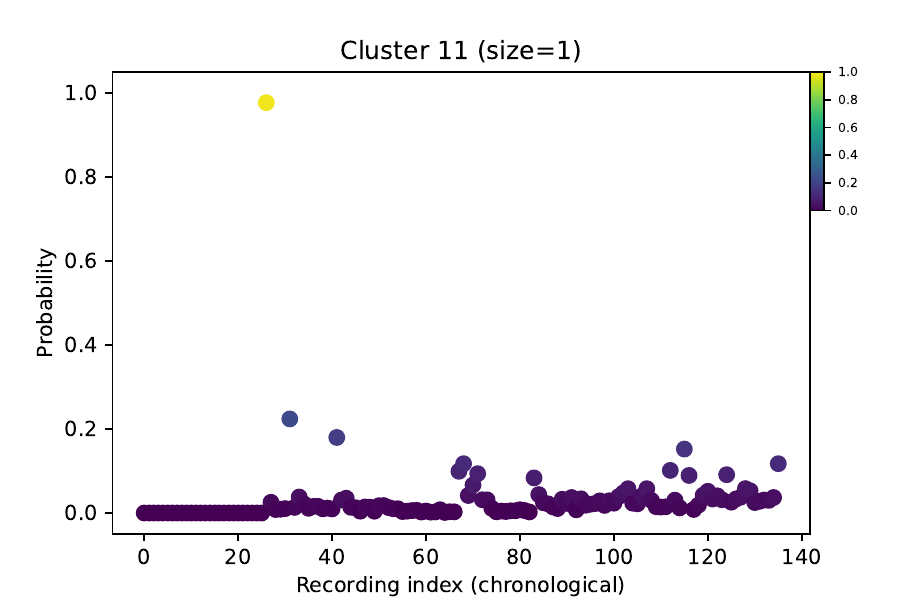}%
\label{fig:c11}}\hfil
\subfloat{\includegraphics[width=0.25\linewidth]{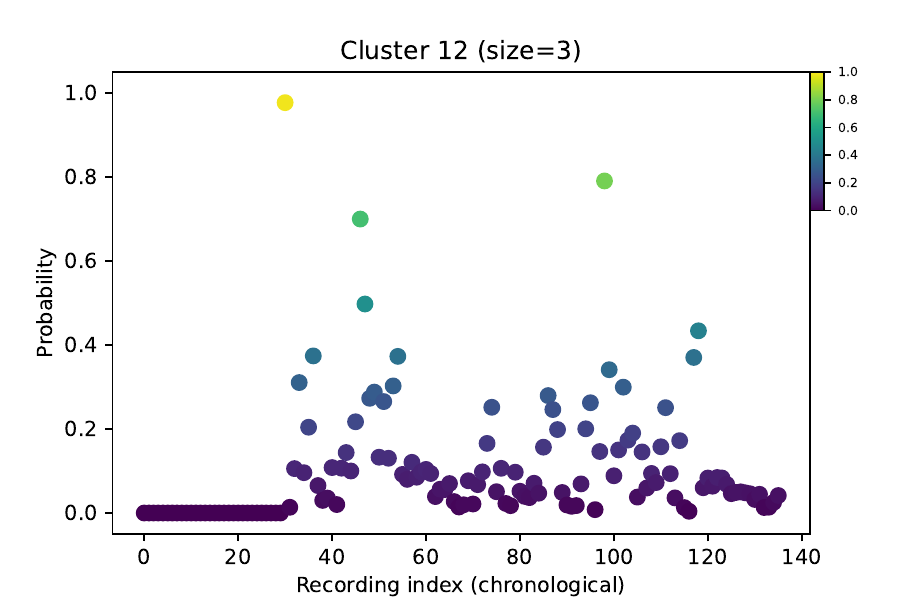}%
\label{fig:c12}}\hfil
\subfloat{\includegraphics[width=0.25\linewidth]{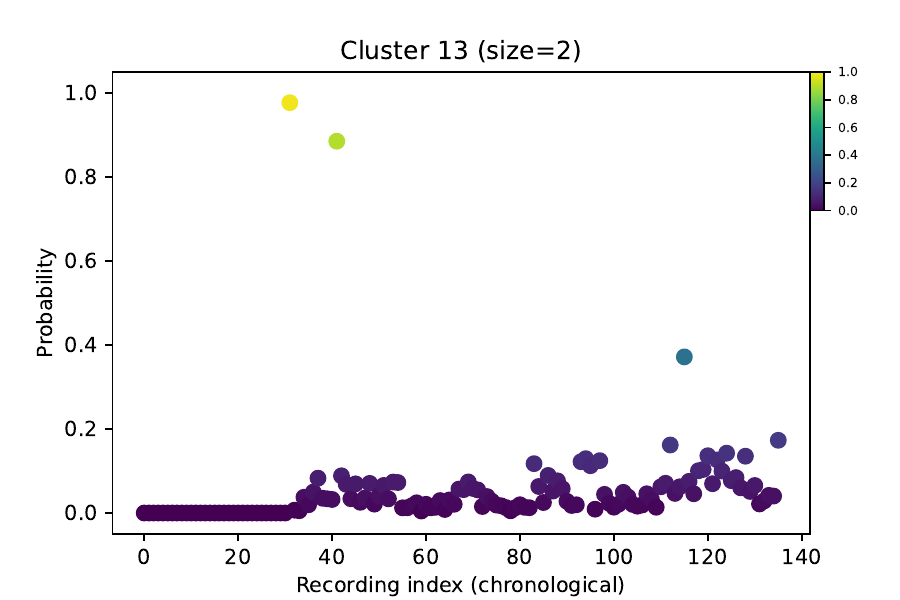}%
\label{fig:c13}}\hfil
\subfloat{\includegraphics[width=0.25\linewidth]{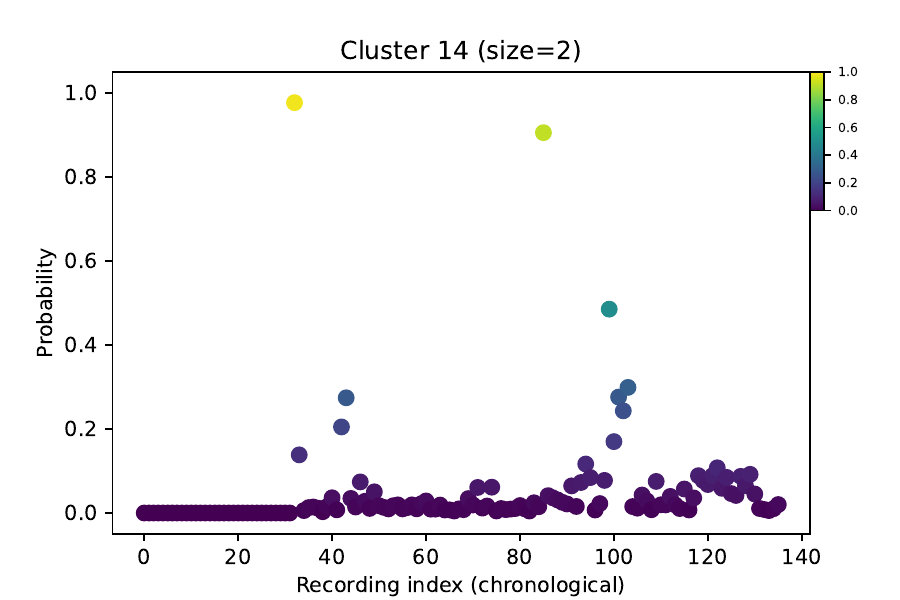}%
\label{fig:c14}}
\vspace{-1.85em}
\subfloat{\includegraphics[width=0.25\linewidth]{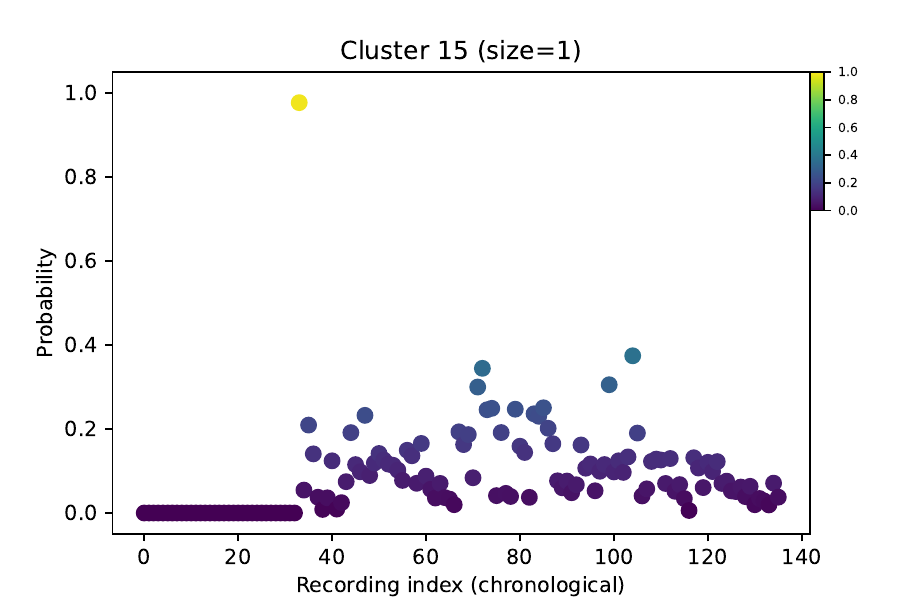}%
\label{fig:c15}}\hfil
\subfloat{\includegraphics[width=0.25\linewidth]{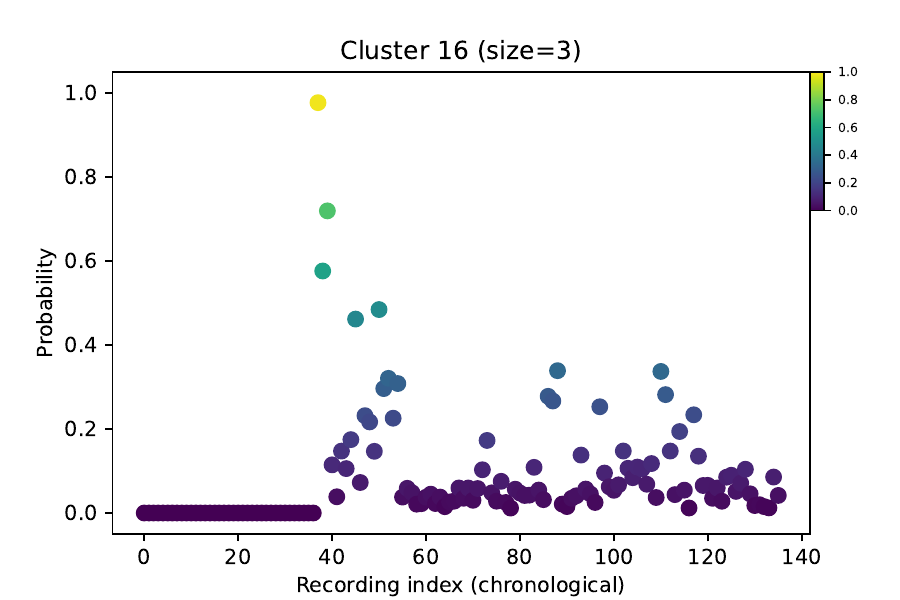}%
\label{fig:c16}}\hfil
\subfloat{\includegraphics[width=0.25\linewidth]{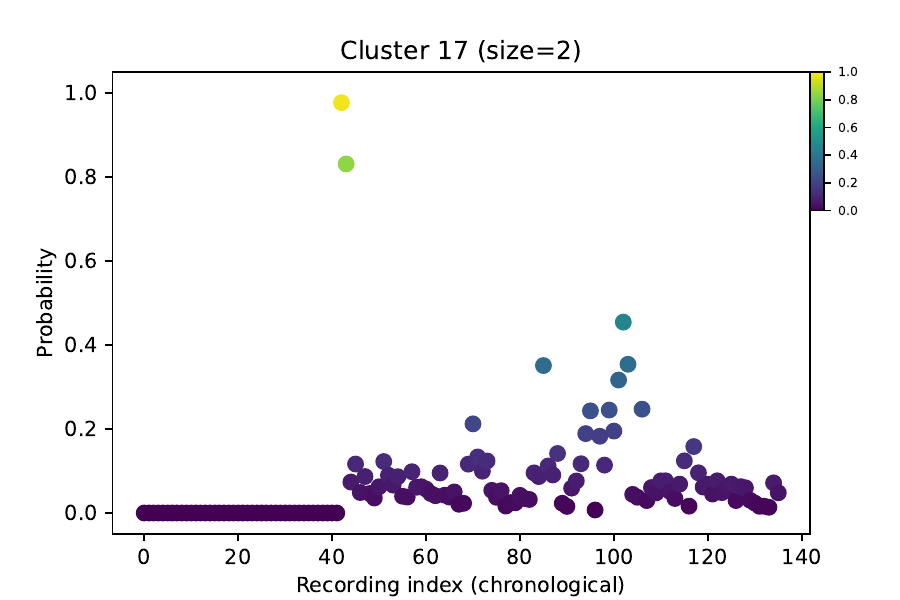}%
\label{fig:c17}}\hfil
\subfloat{\includegraphics[width=0.25\linewidth]{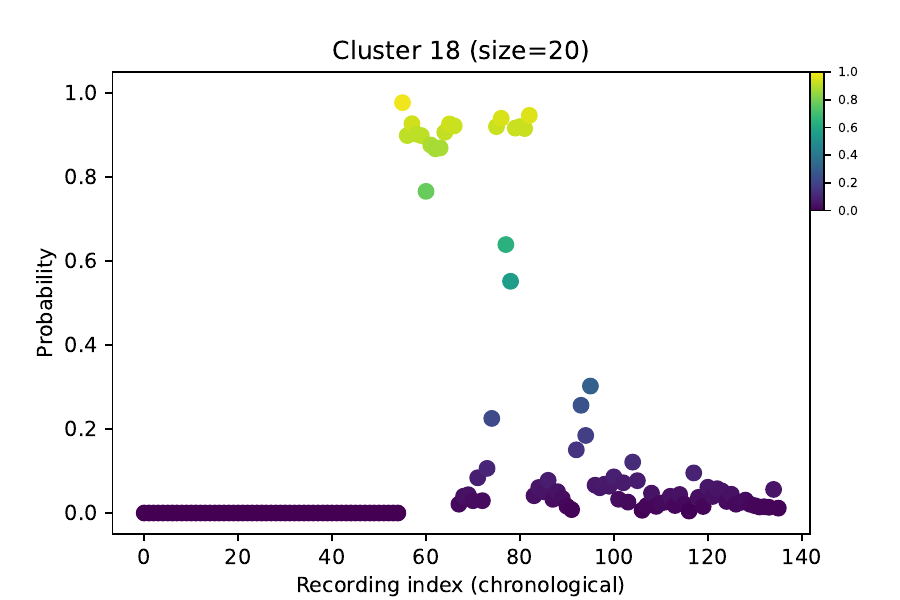}%
\label{fig:c18}}
\vspace{-1.85em}
\subfloat{\includegraphics[width=0.25\linewidth]{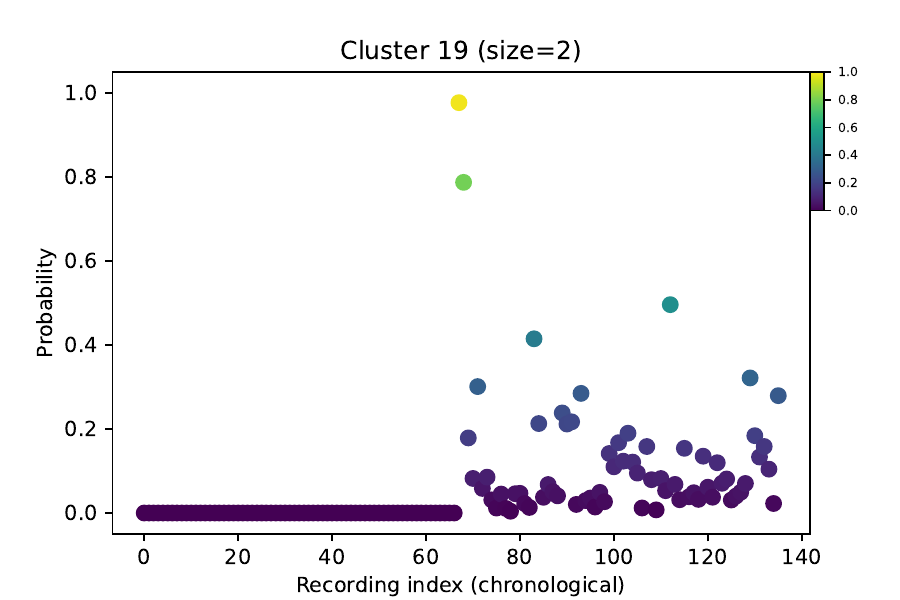}%
\label{fig:c19}}\hfil
\subfloat{\includegraphics[width=0.25\linewidth]{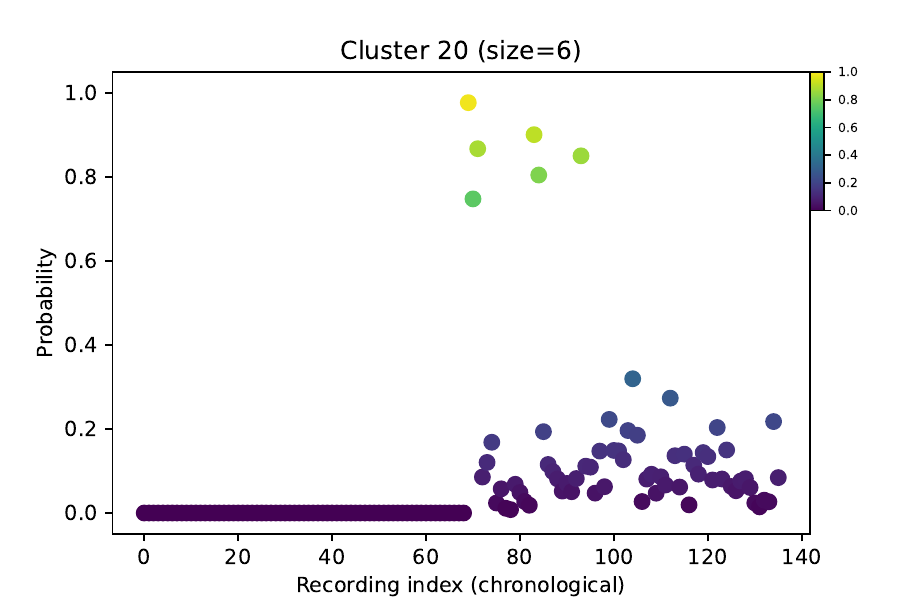}%
\label{fig:c20}}\hfil
\subfloat{\includegraphics[width=0.25\linewidth]{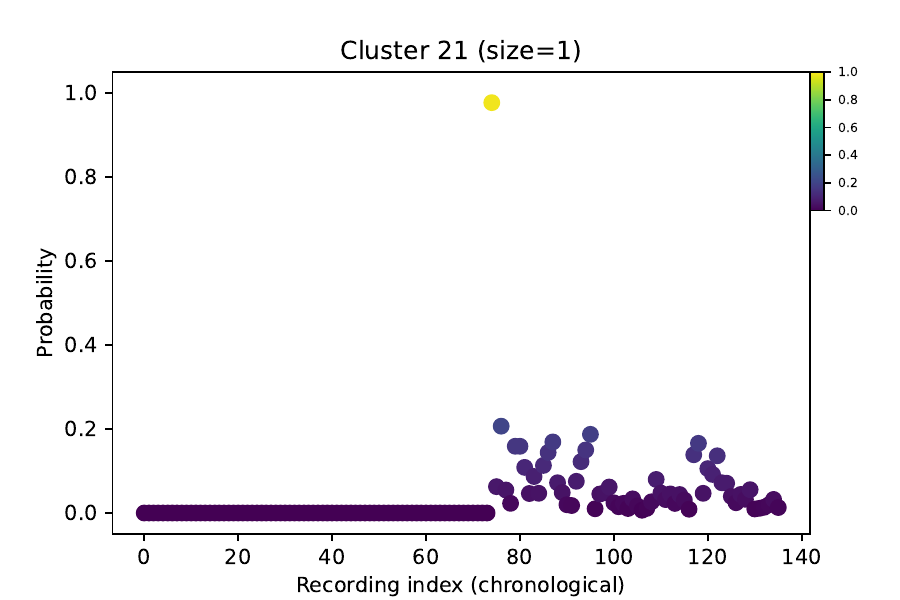}%
\label{fig:c21}}\hfil
\subfloat{\includegraphics[width=0.25\linewidth]{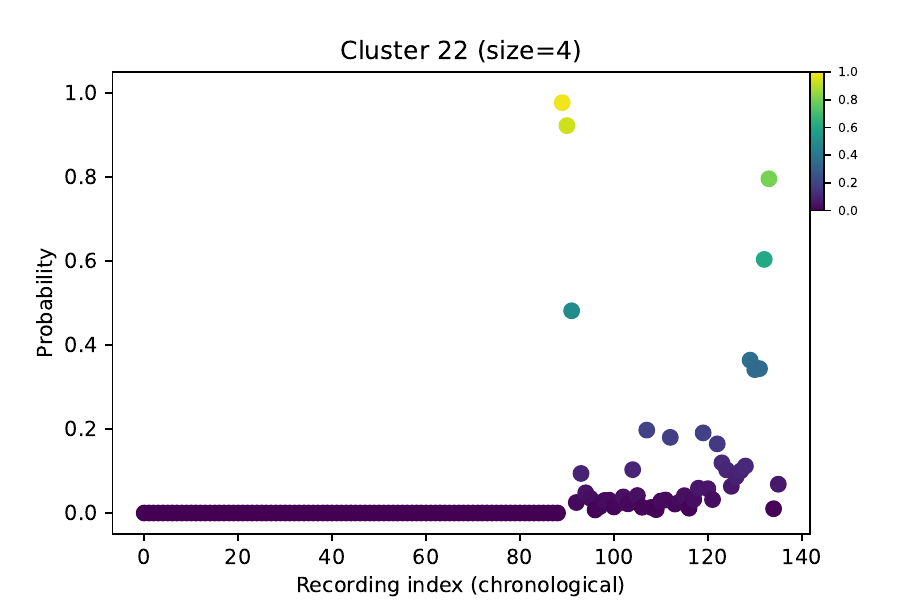}%
\label{fig:c22}}
\vspace{-1.85em}
\subfloat{\includegraphics[width=0.25\linewidth]{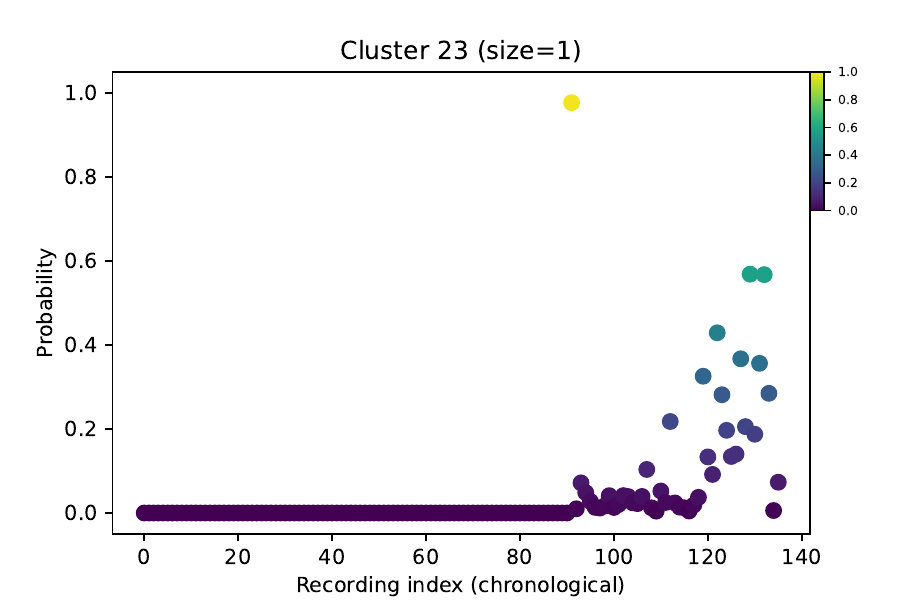}%
\label{fig:c23}}\hfil
\subfloat{\includegraphics[width=0.25\linewidth]{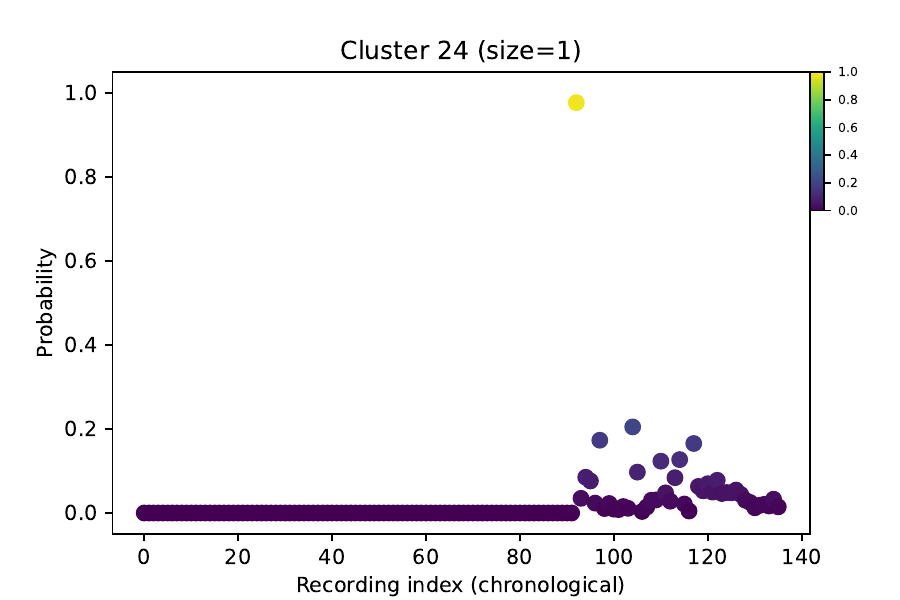}%
\label{fig:c24}}\hfil
\subfloat{\includegraphics[width=0.25\linewidth]{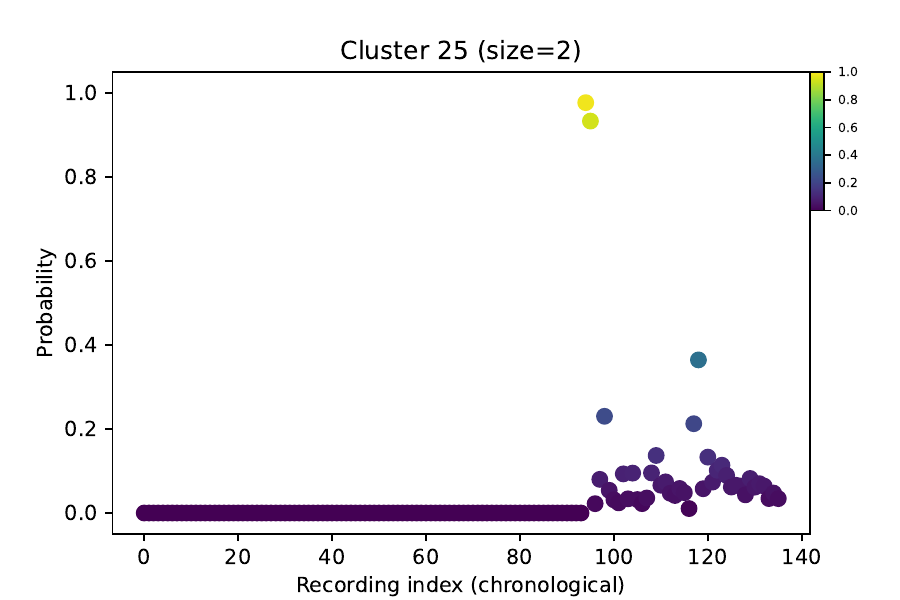}%
\label{fig:c25}}\hfil
\subfloat{\includegraphics[width=0.25\linewidth]{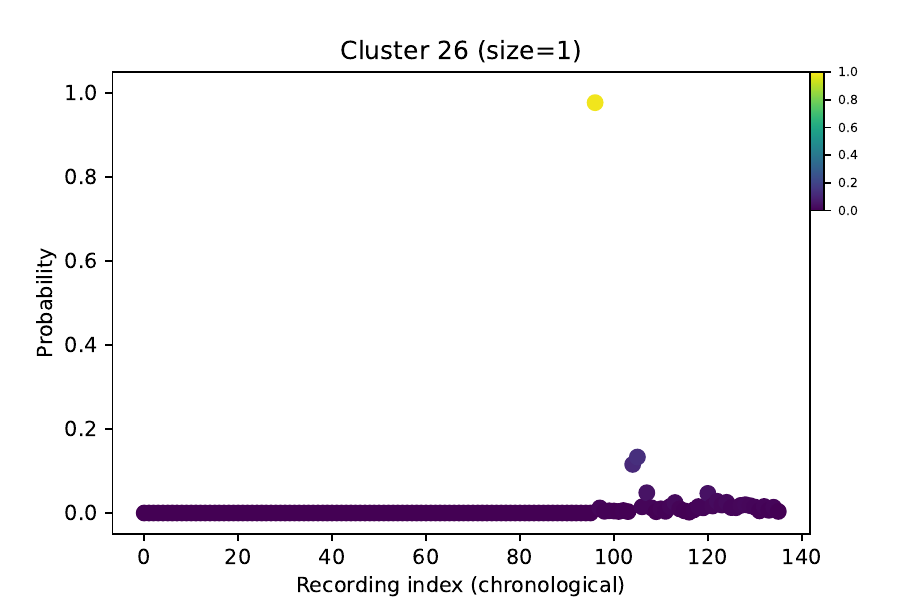}%
\label{fig:c26}}
\vspace{-1.85em}\hfil
\subfloat{\includegraphics[width=0.25\linewidth]{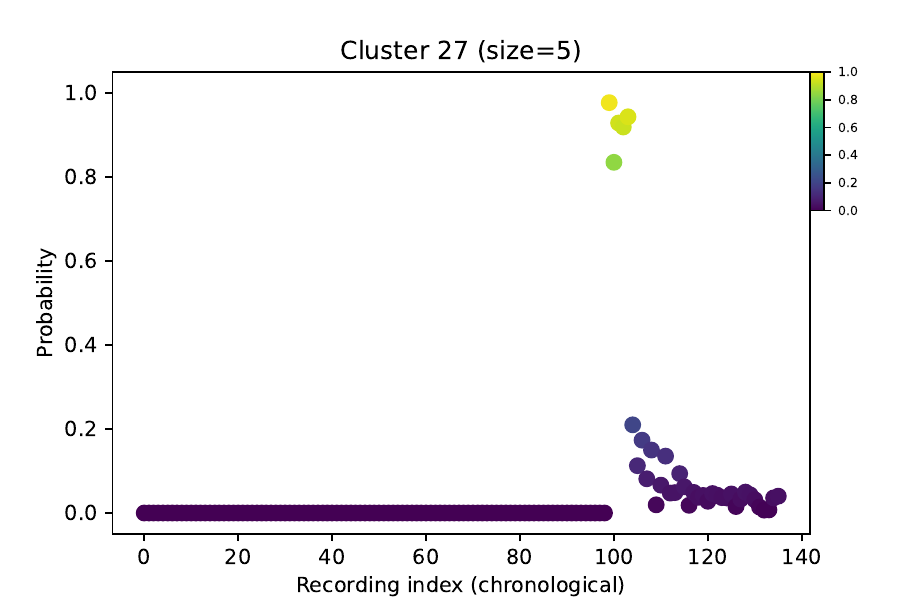}%
\label{fig:c27}}\hfil
\subfloat{\includegraphics[width=0.25\linewidth]{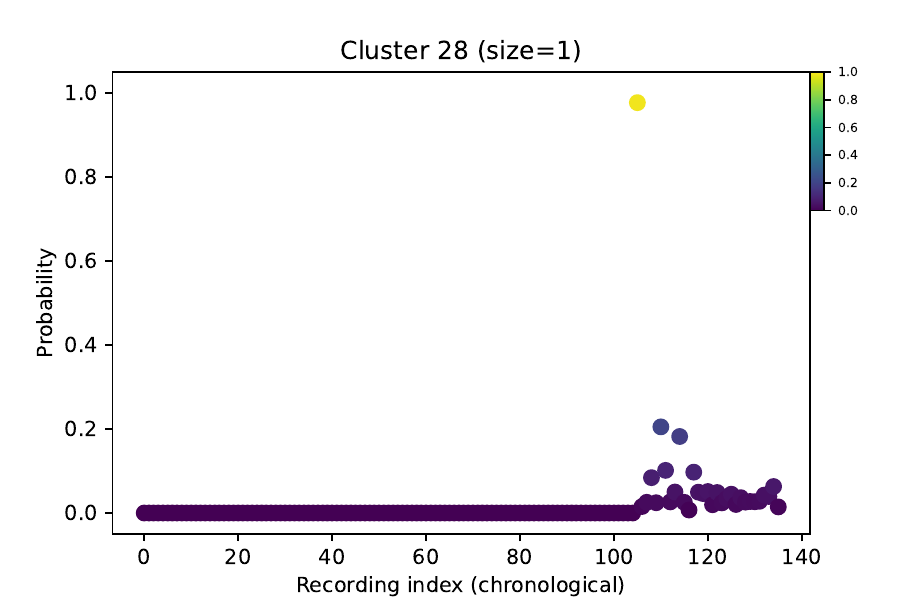}%
\label{fig:c28}}\hfil
\subfloat{\includegraphics[width=0.25\linewidth]{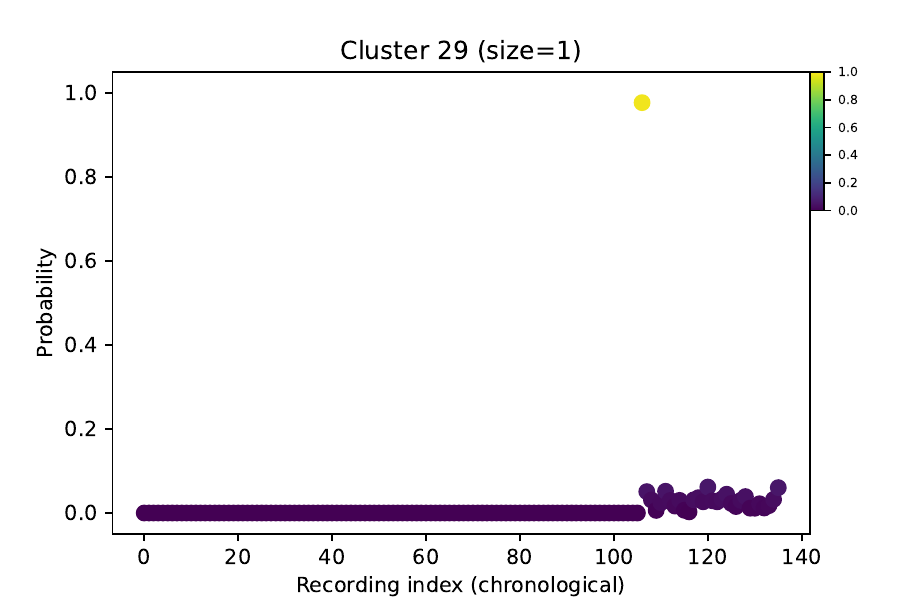}%
\label{fig:c29}}\hfil
\subfloat{\includegraphics[width=0.25\linewidth]{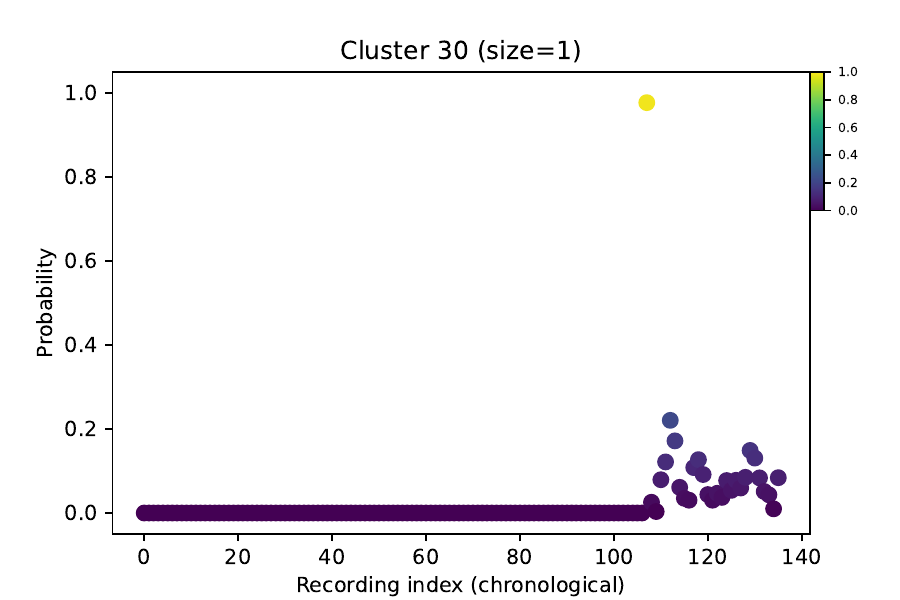}%
\label{fig:c30}}
\vspace{-1.85em}
\subfloat{\includegraphics[width=0.25\linewidth]{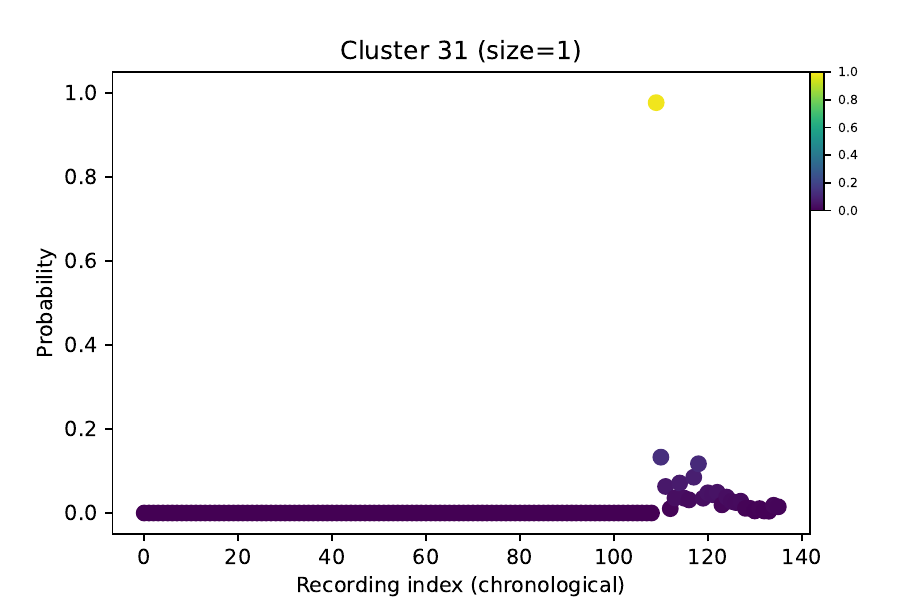}%
\label{fig:c31}}\hfil
\subfloat{\includegraphics[width=0.25\linewidth]{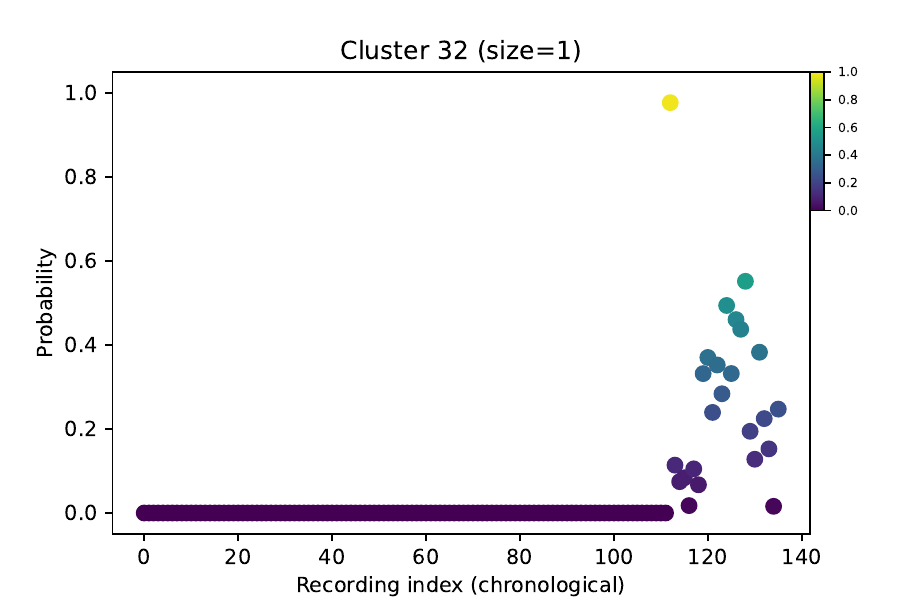}%
\label{fig:c32}}\hfil
\subfloat{\includegraphics[width=0.25\linewidth]{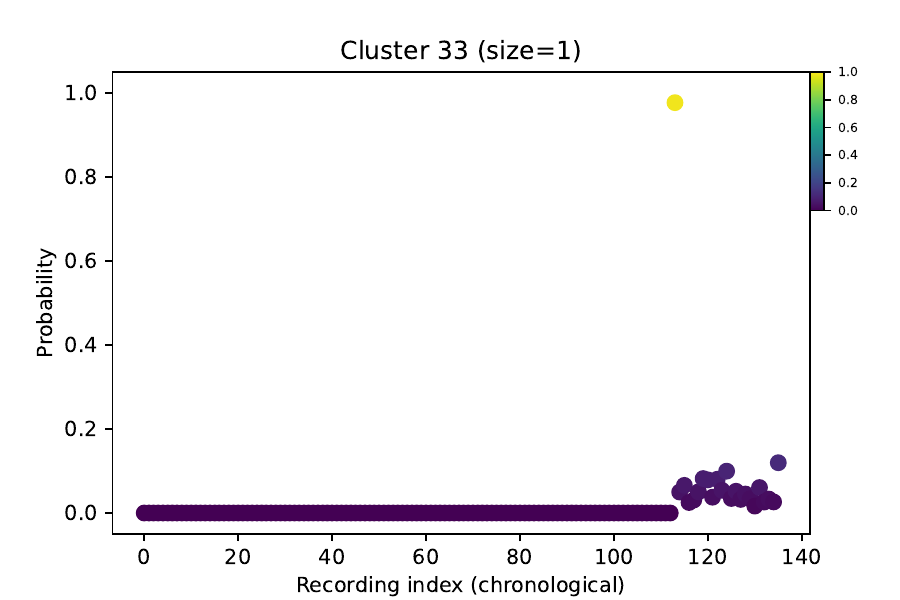}%
\label{fig:c33}}\hfil
\subfloat{\includegraphics[width=0.25\linewidth]{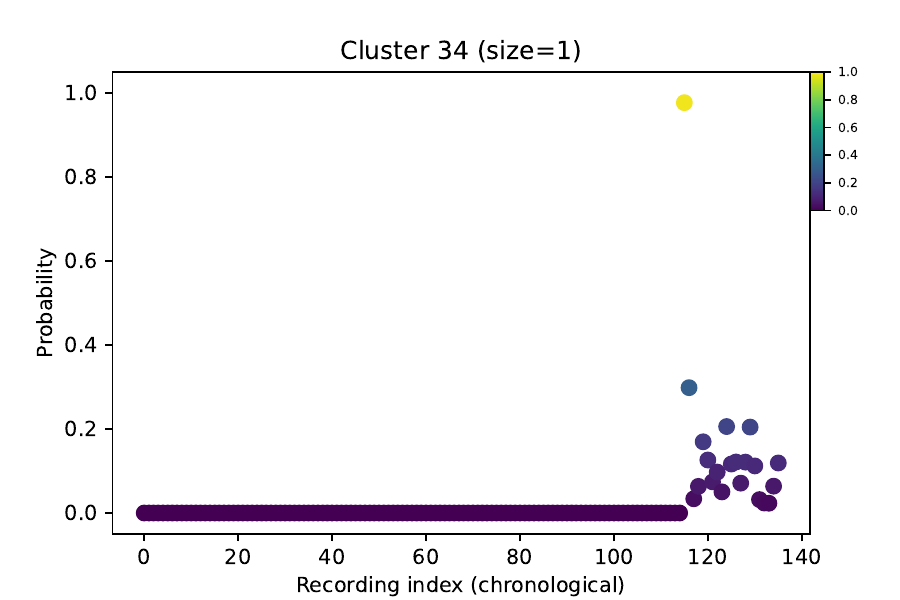}%
\label{fig:c34}}
\vspace{-1.85em}
\subfloat{\includegraphics[width=0.25\linewidth]{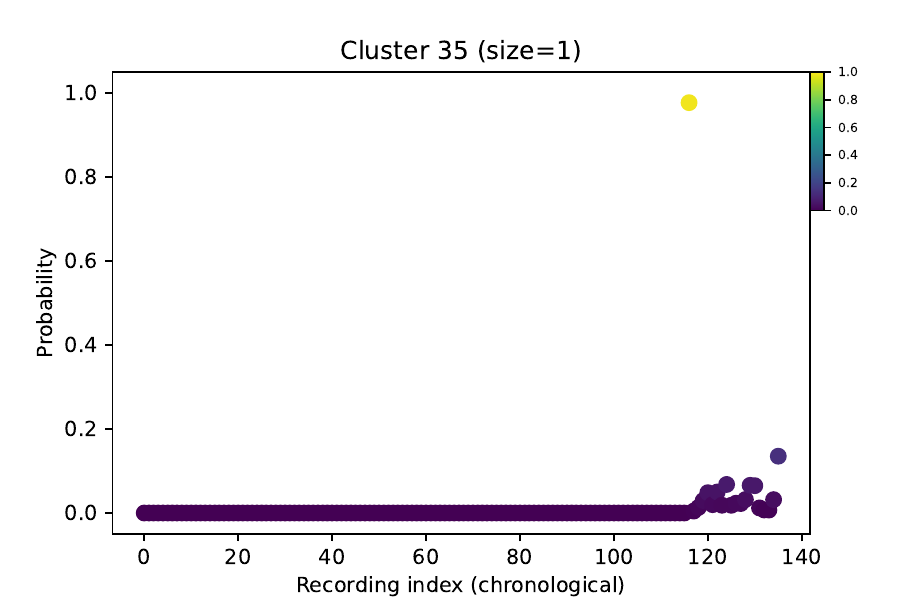}%
\label{fig:c35}}\hfil
\subfloat{\includegraphics[width=0.25\linewidth]{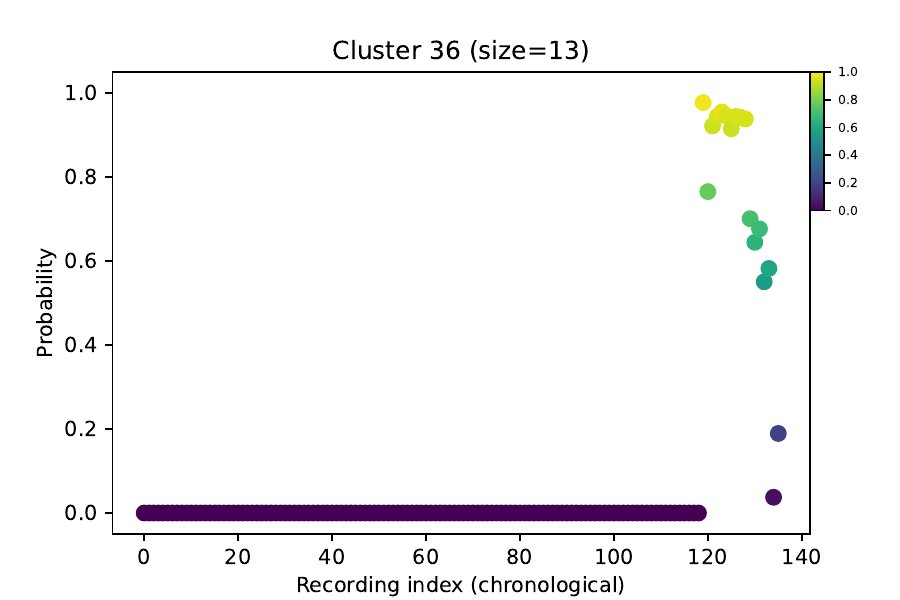}%
\label{fig:c36}}\hfil
\subfloat{\includegraphics[width=0.25\linewidth]{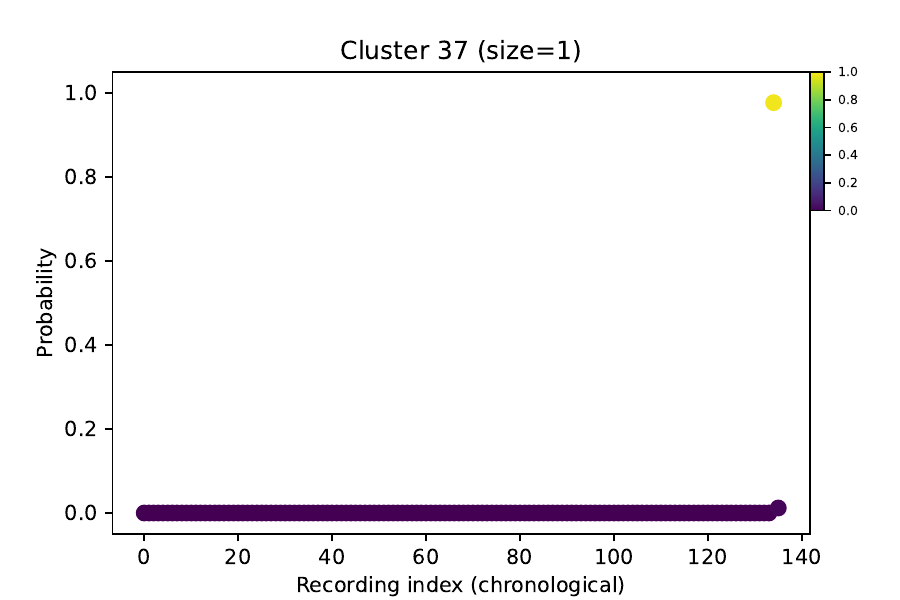}%
\label{fig:c37}}\hfil
\subfloat{\includegraphics[width=0.25\linewidth]{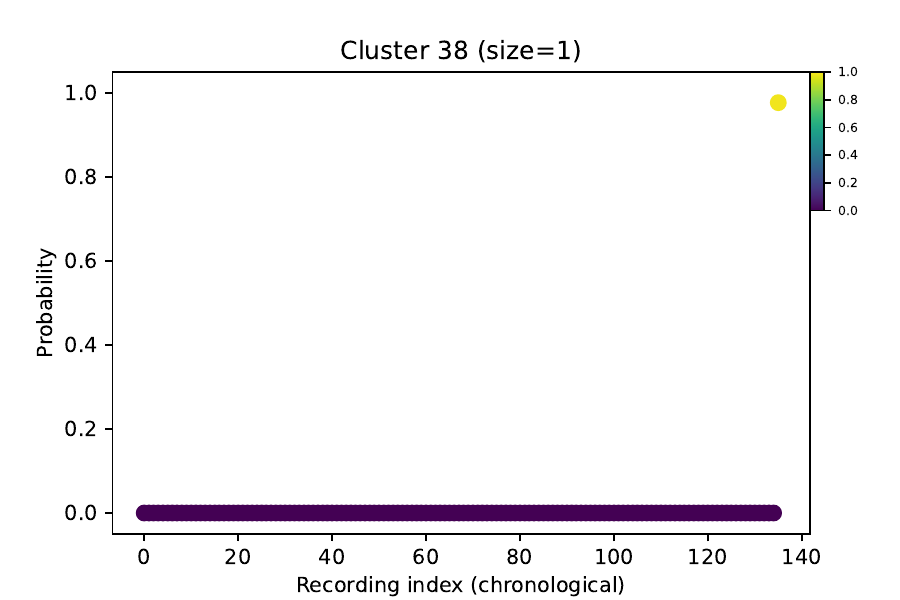}%
\label{fig:c38}}
\vspace{-0.75em}
\caption{Clusters overall detected for common voice recordings based on voice embedding; in addition to \autoref{fig:c026}.}
\label{fig:c1-38m026}
\vspace{-5mm}
\end{figure*}

To benchmark performance, ECAPA–TDNN with DBSCAN, ECAPA–TDNN with cosine similarity, and Resemblyzer’s VoiceEncoder with cosine similarity were evaluated on 136 Common Voice samples (39 speakers, one utterance per speaker). VoiceEncoder with cosine similarity yielded the best results, detecting all 39 speakers with Adjusted Rand Index (ARI) 0.8682, Adjusted Mutual Information (AMI) 0.8729, Homogeneity 0.9250, Completeness 0.9450, and V-measure 0.9349 (\autoref{fig:c026}). \autoref{fig:c1-38m026} extends the analysis shown in \autoref{fig:c026} by presenting the complete cluster formation.

These results validate that the system’s components function reliably, even if full-scale benchmarking on real FNOL data remains future work.

\vspace{2mm}
\section{Conclusion}\label{sec:conclusion}
We present a synthetic multimodal framework for insurance fraud detection, addressing the absence of shareable audio–text datasets constrained by privacy regulations. 
\highlight{This work presents a proof-of-function pipeline that unifies synthetic data generation, speech processing, and fraud detection within a risk scoring architecture. The system is designed to capture structured anomalies, narrative reuse, and repeated voices, while minimizing false positives.
Intrinsic dataset validation and component-level evaluations support the robustness of the approach. 
Future work will expand narrative and speaker diversity, introduce varied adversarial fraud scenarios, and begin validation on anonymized or commercial datasets.
Although this study focuses on text and voice modalities, future milestones will explore the integration of imaging data. These extensions aim to bridge synthetic evaluation with real-world deployment.}

\newpage

\bibliographystyle{IEEEtran}
\bibliography{Ref}

\begin{thebibliography}{10}
\providecommand{\url}[1]{#1}
\csname url@samestyle\endcsname
\providecommand{\newblock}{\relax}
\providecommand{\bibinfo}[2]{#2}
\providecommand{\BIBentrySTDinterwordspacing}{\spaceskip=0pt\relax}
\providecommand{\BIBentryALTinterwordstretchfactor}{4}
\providecommand{\BIBentryALTinterwordspacing}{\spaceskip=\fontdimen2\font plus
\BIBentryALTinterwordstretchfactor\fontdimen3\font minus \fontdimen4\font\relax}
\providecommand{\BIBforeignlanguage}[2]{{%
\expandafter\ifx\csname l@#1\endcsname\relax
\typeout{** WARNING: IEEEtran.bst: No hyphenation pattern has been}%
\typeout{** loaded for the language `#1'. Using the pattern for}%
\typeout{** the default language instead.}%
\else
\language=\csname l@#1\endcsname
\fi
#2}}
\providecommand{\BIBdecl}{\relax}
\BIBdecl

\bibitem{govuk2024fraud}
{Home Office} and {The Rt Hon Lord Hanson of Flint}, ``Major new crackdown on insurance fraud,'' \url{https://www.gov.uk/government/news/major-new-crackdown-on-insurance-fraud}, October 2024, accessed August 2025.

\bibitem{ifb2024report}
{Insurance Fraud Bureau}, ``Ifb’s 2024 annual report has been published,'' \url{https://www.insurancefraudbureau.org/media-centre/ifb-news/2025/ifb-s-2024-annual-report-has-been-published}, 2025, accessed August 2025.

\bibitem{naic2024fraud}
{National Association of Insurance Commissioners}, ``Insurance topics | insurance fraud,'' \url{https://content.naic.org/insurance-topics/insurance-fraud}, 2024, accessed August 2025.

\bibitem{aslam2022insurance}
F.~Aslam, A.~I. Hunjra, Z.~Ftiti, W.~Louhichi, and T.~Shams, ``Insurance fraud detection: Evidence from artificial intelligence and machine learning,'' \emph{Technological Forecasting and Social Change}, 2022.

\bibitem{ali2022financial}
A.~Ali, S.~A. Razak, S.~H. Othman, T.~A.~E. Eisa, A.~Al-Dhaqm, M.~Nasser, T.~Elhassan, H.~Elshafie, and A.~Saif, ``Financial fraud detection based on machine learning: A systematic literature review,'' \emph{IEEE Access}, 2022.

\bibitem{backlund2023detection}
R.~B{\"a}cklund and H.~{\"O}hman, ``Detection of insurance fraud using nlp and ml: A study on three different nlp-techniques for text classification,'' Master's thesis, Lund University, 2023.

\bibitem{devlin2019bert}
J.~Devlin, M.-W. Chang, K.~Lee, and K.~Toutanova, ``Bert: Pre-training of deep bidirectional transformers for language understanding,'' in \emph{Proceedings of the 2019 conference of the North American chapter of the association for computational linguistics: human language technologies, volume 1 (long and short papers)}, 2019, pp. 4171--4186.

\bibitem{liu2024gpt}
X.~Liu, Y.~Zheng, Z.~Du, M.~Ding, Y.~Qian, Z.~Yang, and J.~Tang, ``Gpt understands, too,'' \emph{AI Open}, vol.~5, pp. 208--215, 2024.

\bibitem{yang2019xlnet}
Z.~Yang, Z.~Dai, Y.~Yang, J.~Carbonell, R.~R. Salakhutdinov, and Q.~V. Le, ``Xlnet: Generalized autoregressive pretraining for language understanding,'' \emph{Advances in neural information processing systems}, vol.~32, 2019.

\bibitem{piehl2021classification}
C.~Piehl, ``Classification of transcribed voice recordings: Determining the claim type of recordings submitted by swedish insurance clients,'' Master's thesis, KTH Royal Institute of Technology, 2021.

\bibitem{mendeley2020insurance}
A.~Abdelrahim and M.~Data, ``Insurance claim fraud dataset,'' 2020, \url{https://data.mendeley.com/datasets/992mh7dk9y/1}.

\bibitem{aixblock2024calls}
AIxBlock, ``92k real-world call center scripts (english),'' \url{https://huggingface.co/datasets/AIxBlock/92k-real-world-call-center-scripts-english}, 2024, accessed August 2025.

\bibitem{leal2018telephone}
S.~Leal, A.~Vrij, L.~Warmelink, Z.~Vernham, and R.~P. Fisher, ``You cannot hide your telephone lies: Providing a model statement as an aid to detect deception in insurance telephone calls,'' \emph{Applied Cognitive Psychology}, vol.~32, no.~6, 2018.

\bibitem{bajaj2020fraud}
N.~Bajaj, T.~G. Constance, M.~Rajwadi, J.~Wall, M.~Moniri, C.~Glackin, N.~Cannings, C.~Woodruff, and J.~Laird, ``Fraud detection in telephone conversations for financial services using linguistic features,'' in \emph{Proceedings of Interspeech}, 2020.

\bibitem{banulescu2024practical}
D.~Banulescu-Radu and M.~Yankol-Schalck, ``Practical guideline to efficiently detect insurance fraud in the era of machine learning: A household insurance case,'' \emph{Journal of Risk and Insurance}, vol.~91, no.~4, pp. 867--913, 2024.

\bibitem{chang2022design}
J.-W. Chang, N.~Yen, and J.~C. Hung, ``Design of a nlp-empowered finance fraud awareness model: the anti-fraud chatbot for fraud detection and fraud classification as an instance,'' \emph{Journal of Ambient Intelligence and Humanized Computing}, vol.~13, no.~10, pp. 4663--4679, 2022.

\bibitem{chang2023design}
------, ``Design of a nlp-empowered finance fraud awareness model: the anti-fraud chatbot for fraud detection and fraud classification as an instance,'' \emph{Journal of Ambient Intelligence and Humanized Computing}, 2023.

\bibitem{dimri2024enhancing}
A.~Dimri, S.~Yerramilli, P.~Lee, S.~Afra, and A.~Jakubowski, ``Enhancing claims handling processes with insurance based language models,'' in \emph{Proceedings of the AAAI Workshop on AI in Insurance}, 2024.

\bibitem{gangani2023ai}
C.~M. Gangani, ``Ai in insurance: Enhancing fraud detection and risk assessment,'' \emph{International Journal of Advanced Computer Science and Applications}, 2023.

\bibitem{tarra2024ai}
V.~K. Tarra, ``Ai in fraud detection: Leveraging machine learning to combat insurance fraud,'' \emph{International Journal of Innovative Technology and Exploring Engineering}, 2024.

\bibitem{perumal2023innovative}
R.~A. Perumal, ``Innovative applications of ai and machine learning in fraud detection for insurance claims,'' \emph{Journal of Risk and Financial Management}, 2023.

\bibitem{banulescu2023practical}
D.~Banulescu‐Radu and M.~Yankol‐Schalck, ``Practical guideline to efficiently detect insurance fraud in the era of machine learning: A household insurance case,'' \emph{Journal of Risk Finance}, 2023.

\bibitem{kumar2022detecting}
P.~K. Kumar, S.~Ray, L.~Kumarasankaralingam, A.~Ramamoorthy, P.~Kumar, and A.~Dutta, ``Detecting fraud calls vis-à-vis natural language processing,'' in \emph{Proceedings of the International Conference on Computer Communication and Informatics}, 2022.

\bibitem{gupta2024detection}
A.~Gupta, ``Detection of spam and fraudulent calls using natural language processing model,'' \emph{International Journal of Computer Applications}, 2024.

\bibitem{nie2025multimodal}
H.~Nie, Z.~Long, Z.~Fang, and L.~Gao, ``Multimodal detection framework for financial fraud integrating llms and interpretable machine learning,'' \emph{Journal of Data and Information Science}, vol.~10, no.~4, pp. 1--25, 2025.

\bibitem{dupreez2023healthcare}
A.~du~Preez, S.~Bhattachary, P.~Beling, and E.~Bowen, ``Fraud detection in healthcare claims using machine learning: A systematic review,'' \emph{Health Informatics Journal}, 2023.

\bibitem{yang2023auto}
J.~Yang, K.~Chen, K.~Ding, C.~Na, and M.~Wang, ``Auto insurance fraud detection with multimodal learning,'' \emph{Data Intelligence}, vol.~5, no.~2, pp. 388--412, 2023.

\bibitem{asgarian2023autofraudnet}
A.~Asgarian, R.~Saha, D.~Jakubovitz, and J.~Peyre, ``Autofraudnet: A multimodal network to detect fraud in the auto insurance industry,'' \emph{arXiv preprint arXiv:2301.07526}, 2023.

\bibitem{swissre2025genai}
F.~Maurer and V.~Plantard, ``How generative ai is transforming insurance claims: Inside swiss re’s claimsgenai,'' 2025, \url{https://www.swissre.com/risk-knowledge/advancing-societal-benefits-digitalisation/how-generative-ai-is-transforming-insurance-claims-claimsgenai.html}.

\bibitem{xenoss2025fraud}
Xenoss, ``Enterprise fraud detection systems | custom ai solutions,'' 2025, \url{https://xenoss.io/solutions/fraud-detection}.

\bibitem{moodys2025fraud}
M.~Analytics, ``Fraud risk management | moody’s data \& analytics solutions,'' 2025, \url{https://www.moodys.com/web/en/us/kyc/solutions/fraud-prevention.html}.

\bibitem{mutemi2023ecommerce}
A.~Mutemi and F.~Bacao, ``E-commerce fraud detection based on machine learning techniques: Systematic literature review,'' \emph{Journal of Information Security and Applications}, 2023.

\bibitem{boulieris2024fraudnlp}
P.~Boulieris, J.~Pavlopoulos, A.~Xenos, and V.~Vassalos, ``Fraud detection with natural language processing,'' \emph{Machine Learning}, 2024, \url{https://github.com/pboulieris/FraudNLP}.

\bibitem{nirab2021insuranceclaim}
Nirab, ``Insurance-claim-fraud-detection (github repository),'' 2021, \url{https://github.com/nirab25/Insurance-Claim-Fraud-Detection}.

\bibitem{prapra2023insurancefraud}
P.~S. Prakash, ``Insurance claims fraud detection model (github repository),'' 2023, \url{https://github.com/PrajwalSuryaPrakash/Insurance-Claims-Fraud-Detection-Model}.

\bibitem{manoj2023predictive}
M.~Gaikwad, ``Predictive analysis and fraud detection for insurance claims (github pages),'' 2023, \url{https://manojgaikwad13.github.io/Predictive-Analysis-and-Fraud-Detection-for-Insurance-Claims/}.

\bibitem{cheuk2023frauddetector}
Y.~Cheuk, ``Fraud detector (github repository),'' 2023, \url{https://github.com/yingcheuk/fraud-detector}.

\bibitem{deep_learning_nlp_guide}
{DeepLearning.AI}, ``A complete guide to natural language processing,'' \url{https://www.deeplearning.ai/resources/natural-language-processing/}, 2025, accessed: 17-July-2025.

\bibitem{ONS2025Families}
\BIBentryALTinterwordspacing
{Office for National Statistics}, ``Families and households,'' 2025, accessed: 2025-08-29. [Online]. Available: \url{https://www.ons.gov.uk/peoplepopulationandcommunity/birthsdeathsandmarriages/families/datasets/familiesandhouseholdsfamiliesandhouseholds}
\BIBentrySTDinterwordspacing

\bibitem{wolf-etal-2020-transformers}
\BIBentryALTinterwordspacing
T.~Wolf, L.~Debut, V.~Sanh, J.~Chaumond, C.~Delangue, A.~Moi, P.~Cistac, T.~Rault, R.~Louf, M.~Funtowicz, J.~Davison, S.~Shleifer, P.~von Platen, C.~Ma, Y.~Jernite, J.~Plu, C.~Xu, T.~L. Scao, S.~Gugger, M.~Drame, Q.~Lhoest, and A.~M. Rush, ``Transformers: State-of-the-art natural language processing,'' in \emph{Proceedings of the 2020 Conference on Empirical Methods in Natural Language Processing: System Demonstrations}.\hskip 1em plus 0.5em minus 0.4em\relax Online: Association for Computational Linguistics, Oct. 2020, pp. 38--45. [Online]. Available: \url{https://www.aclweb.org/anthology/2020.emnlp-demos.6}
\BIBentrySTDinterwordspacing

\bibitem{radford2019language}
A.~Radford, J.~Wu, R.~Child, D.~Luan, D.~Amodei, and I.~Sutskever, ``Language models are unsupervised multitask learners,'' 2019, openAI.

\bibitem{coqui2023xtts}
C.~AI, ``Coqui tts/xtts: A deep learning toolkit for high-quality text-to-speech,'' \url{https://github.com/coqui-ai/TTS}, 2023.

\bibitem{casanova2024xtts}
E.~Casanova, K.~Davis, E.~G{\"o}lge, G.~G{\"o}knar, I.~Gulea, L.~Hart, A.~Aljafari, J.~Meyer, R.~Morais, S.~Olayemi \emph{et~al.}, ``Xtts: a massively multilingual zero-shot text-to-speech model,'' \emph{arXiv preprint arXiv:2406.04904}, 2024.

\bibitem{gtts2024}
D.~Pellegrini, ``gtts: Google text-to-speech python library,'' \url{https://gtts.readthedocs.io/}, 2024.

\bibitem{bain2023whisperx}
M.~Bain, J.~Huh, T.~Han, and A.~Zisserman, ``Whisperx: Time-accurate speech transcription of long-form audio,'' \emph{INTERSPEECH 2023}, 2023.

\bibitem{roberta_large_ner_english}
J.-B. Polle, ``Jean-baptiste/roberta-large-ner-english,'' \url{https://huggingface.co/Jean-Baptiste/roberta-large-ner-english}, 2021, fine-tuned RoBERTa model for Named Entity Recognition on CoNLL-2003 dataset.

\bibitem{sentence-transformers-all-minilm-l6-v2}
\BIBentryALTinterwordspacing
Sentence-Transformers, ``all-minilm-l6-v2,'' accessed: 2025-09-05. [Online]. Available: \url{https://huggingface.co/sentence-transformers/all-MiniLM-L6-v2}
\BIBentrySTDinterwordspacing

\bibitem{douze2024faiss}
M.~Douze, A.~Guzhva, C.~Deng, J.~Johnson, G.~Szilvasy, P.-E. Mazaré, M.~Lomeli, L.~Hosseini, and H.~Jégou, ``The faiss library,'' \emph{arXiv}, 2024.

\bibitem{johnson2019billion}
J.~Johnson, M.~Douze, and H.~J{\'e}gou, ``Billion-scale similarity search with {GPUs},'' \emph{IEEE Transactions on Big Data}, vol.~7, no.~3, pp. 535--547, 2019.

\bibitem{lewis2020retrieval}
P.~Lewis, E.~Perez, A.~Piktus, F.~Petroni, V.~Karpukhin, N.~Goyal, H.~K{\"u}ttler, M.~Lewis, W.-t. Yih, T.~Rockt{\"a}schel \emph{et~al.}, ``Retrieval-augmented generation for knowledge-intensive nlp tasks,'' \emph{Advances in neural information processing systems}, vol.~33, pp. 9459--9474, 2020.

\bibitem{langchain_repo}
{LangChain}, ``Langchain,'' \url{https://github.com/langchain-ai/langchain}, 2025, accessed: 17-July-2025.

\bibitem{resemblyzer2020}
H.~W. Corentin~Jemine \emph{et~al.}, ``Resemblyzer: Voice embeddings for speaker similarity,'' \url{https://github.com/resemble-ai/Resemblyzer}, 2020.

\bibitem{scikit-learn}
F.~Pedregosa, G.~Varoquaux, A.~Gramfort, V.~Michel, B.~Thirion, O.~Grisel, M.~Blondel, P.~Prettenhofer, R.~Weiss, V.~Dubourg, J.~Vanderplas, A.~Passos, D.~Cournapeau, M.~Brucher, M.~Perrot, and E.~Duchesnay, ``Scikit-learn: Machine learning in {P}ython,'' \emph{Journal of Machine Learning Research}, vol.~12, pp. 2825--2830, 2011.

\bibitem{commonvoice}
Mozilla, ``Common voice,'' \url{https://commonvoice.mozilla.org/pcm/datasets}, 2024, accessed: 2025-09-01.

\end{thebibliography}
\balance

\end{document}